\documentclass{article} 
\usepackage{iclr2024_conference,times}

\usepackage{amsmath,amsfonts,bm}

\def\eqref#1{equation~\ref{#1}}

\def\1{\bm{1}}

\DeclareMathAlphabet{\mathsfit}{\encodingdefault}{\sfdefault}{m}{sl}
\SetMathAlphabet{\mathsfit}{bold}{\encodingdefault}{\sfdefault}{bx}{n}

\usepackage{url}

\usepackage[pagebackref,breaklinks,colorlinks]{hyperref}

\usepackage[utf8]{inputenc} 
\usepackage[T1]{fontenc}    

\usepackage{booktabs}
\usepackage{graphicx}
\usepackage{bbm}
\usepackage{caption}
\usepackage{wrapfig}

\usepackage{pifont}

\usepackage{cite}
\usepackage{amsmath,amssymb,amsfonts}
\usepackage{textcomp}
\usepackage{subcaption}
\usepackage{tabularx}
\usepackage{booktabs}       
\usepackage[normalem]{ulem}
\useunder{\uline}{\ul}{}
\usepackage{bm}
\usepackage{color}
\usepackage{xcolor}
\usepackage{algorithm}
\usepackage{algorithmicx}
\usepackage{algpseudocode}
\usepackage{multirow}
\usepackage{xspace}

\usepackage[utf8]{inputenc}
\usepackage[T1]{fontenc}
\usepackage[margin=1in]{geometry}
\usepackage{times}

\usepackage{algpseudocode}
\usepackage{enumitem}
\usepackage{lipsum}
\usepackage{hyperref}

\usepackage{booktabs}
\usepackage{colortbl}
\usepackage{xcolor}
\usepackage{multirow}

\usepackage{tcolorbox}
\tcbuselibrary{breakable}

\definecolor{lightgray}{gray}{0.9}
\definecolor{Gray}{gray}{0.9}

\usepackage{adjustbox}

\hypersetup{
    colorlinks=true,
    linkcolor=blue,
    filecolor=magenta,      
    urlcolor=cyan,
    citecolor=blue,
}

\newcommand{\symboletongyi}{\raisebox{0pt}{~\includegraphics[scale=0.05]{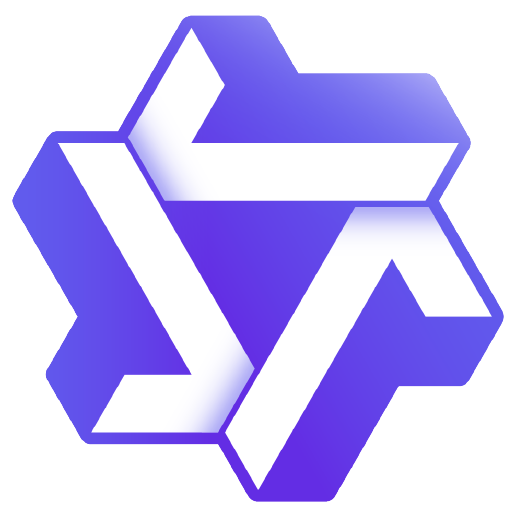}}~}

\usepackage{threeparttable}
\usepackage{newtxtext}
\usepackage[capitalize]{cleveref}
\crefname{section}{Sec.}{Secs.}
\Crefname{section}{Section}{Sections}
\Crefname{table}{Table}{Tables}
\crefname{table}{Tab.}{Tabs.}

\definecolor{myblue}{rgb}{0.2,0.2,0.6}
\definecolor{demphcolor}{RGB}{144, 144, 144}
\definecolor{mygray}{gray}{0.4}
\definecolor{lightgray}{rgb}{0.9, 0.9, 0.9}
\definecolor{deepgreen}{RGB}{0,100,0}

\hypersetup{%
  citecolor=teal
}
\hypersetup{linkcolor = black}
\newcommand{\modelname}{QwenLong-L1.5\xspace}

\title{QwenLong-L1.5: Post-Training Recipe for Long-Context Reasoning and Memory Management}

\usepackage[symbol]{footmisc}
\iclrfinalarxiv 
\begin{document}

\author{
}

\maketitle

\begin{center}
   \centering
   \vspace{-16mm}
   \textbf{Weizhou Shen\footnote{Equal contribution}\qquad Ziyi Yang\footnotemark[1] \qquad Chenliang Li\footnotemark[1] \qquad Zhiyuan Lu\qquad Miao Peng \qquad \\ Huashan Sun \qquad Yingcheng Shi \qquad Shengyi Liao \qquad Shaopeng Lai \qquad Bo Zhang \\ \qquad Dayiheng Liu  \qquad Fei Huang \qquad Jingren Zhou \qquad Ming Yan\footnote{Corresponding author}} \\
    {Tongyi Lab\symboletongyi, Alibaba Group}\\
    {\tt\small \{shenweizhou.swz, ym119608\}@alibaba-inc.com} 
   
   \url{https://github.com/Tongyi-Zhiwen/Qwen-Doc}
\end{center}

\begin{abstract}
We introduce \modelname, a model that achieves superior long-context reasoning capabilities through systematic post-training innovations. The key technical breakthroughs of \modelname are as follows: (1) \textbf{Long-Context Data Synthesis Pipeline}: We develop a systematic synthesis framework that generates challenging reasoning tasks requiring multi-hop grounding over globally distributed evidence. By deconstructing documents into atomic facts and their underlying relationships, and then programmatically composing verifiable reasoning questions, our approach creates high-quality training data at scale, moving substantially beyond simple retrieval tasks to enable genuine long-range reasoning capabilities. (2) \textbf{Stabilized Reinforcement Learning for Long-Context Training}: To overcome the critical instability in long-context RL, we introduce task-balanced sampling with task-specific advantage estimation to mitigate reward bias, and propose Adaptive Entropy-Controlled Policy Optimization (AEPO) that dynamically regulates exploration-exploitation trade-offs. These innovations enable stable training on sequences of progressively increasing length. (3) \textbf{Memory-Augmented Architecture for Ultra-Long Contexts}: Recognizing that even extended context windows cannot accommodate arbitrarily long sequences, we develop a memory management framework with multi-stage fusion RL training that seamlessly integrates single-pass reasoning with iterative memory-based processing for tasks exceeding 4M tokens. Based on Qwen3-30B-A3B-Thinking, \modelname achieves performance comparable to GPT-5 and Gemini-2.5-Pro on long-context reasoning benchmarks, surpassing its baseline by 9.90 points on average. On ultra-long tasks (1M$\sim$4M tokens), \modelname's memory-agent framework yields a 9.48-point gain over the agent baseline. Additionally, the acquired long-context reasoning ability translates to enhanced performance in general domains like scientific reasoning, memory tool using, and extended dialogue.

\end{abstract}
\begin{figure}[H]
    \centering
    \vspace{-0.2cm}
    \includegraphics[width=0.999\textwidth]{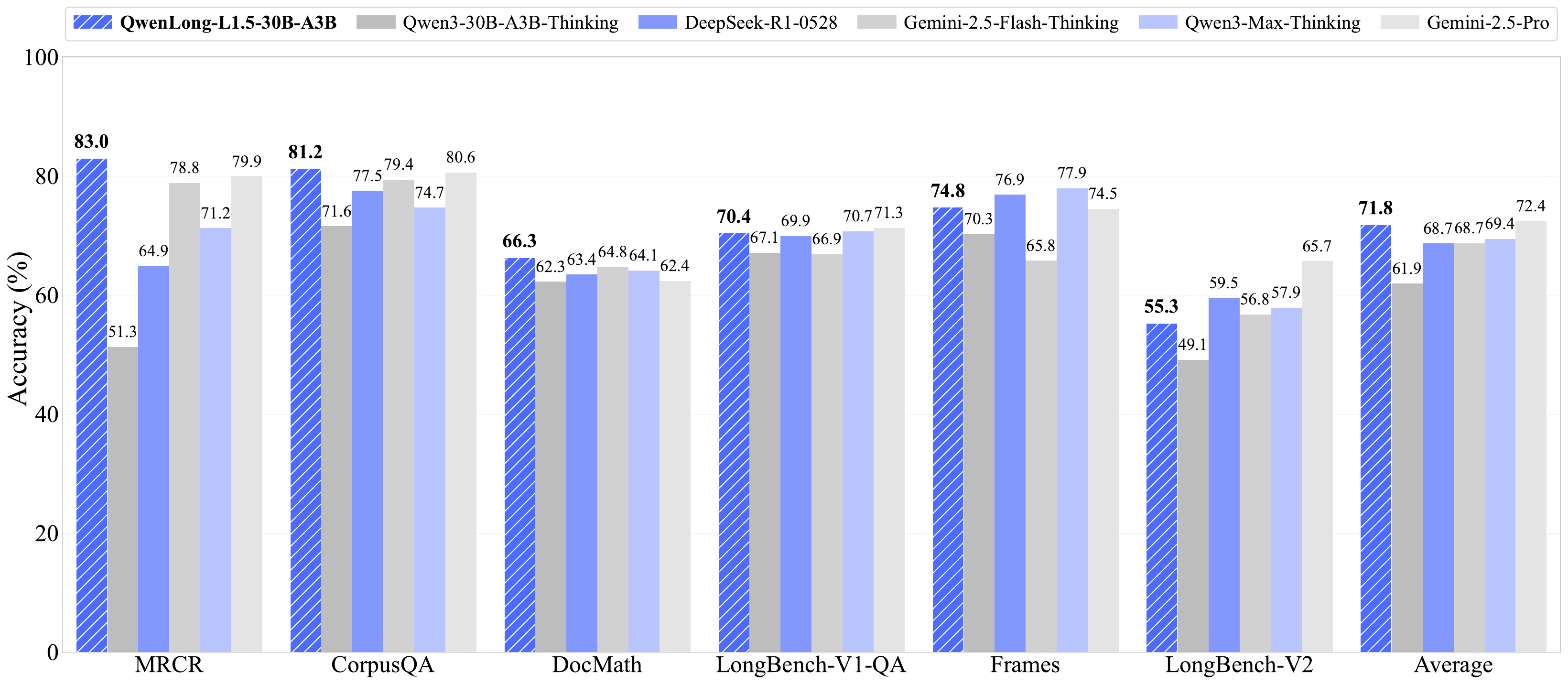}
 \caption{Overall results of {\modelname} across six long-context reasoning benchmarks. Starting from Qwen3-30B-A3B-Thinking, {\modelname}-30B-A3B achieves an average gain of 9.9 points, surpassing DeepSeek-R1-0528, Gemin2.5-Flash-Thinking, Qwen3-Max-Thinking, and comparable to Gemini-2.5-Pro.}
\label{fig:overall_results}
\end{figure}


\section{Introduction}

Long-context reasoning is a critical capability for modern Large Language Models (LLMs), driving advancements in single-pass reasoning~\citep{longbench_v2,zhao2024docmath,krishna2024fact,wu2024longmemeval} and multi-turn autonomous agent systems~\citep{mialon2023gaiabenchmarkgeneralai,patil2025bfcl,barres2025tau2benchevaluatingconversationalagents} by enabling models to integrate key information across a global scope and execute complex multi-hop inference.
Despite its importance, the majority of research on context extension has primarily focused on pre- and mid-training~\citep{qwen3_technical_report, team2025kimi, gemini25} or on architectural innovations~\citep{kimiteam2025kimilinearexpressiveefficient,qiu2025gatedattentionlargelanguage}. A critical gap remains in the post-training stage: the absence of mature, end-to-end systems for long-context reasoning. Specifically, the field lacks a comprehensive post-training recipe that provides: (1) a scalable pipeline for synthesizing challenging, high-value long-context reasoning data; (2) RL-based methods tailored to the nuances of long-context reasoning; and (3) agent architectures designed to operate on information streams that exceed the context capacity.

In this report, we introduce \modelname, a long-context reasoning model built upon Qwen3-30B-A3B-Thinking~\citep{qwen3_technical_report}, augmented with memory mechanisms to process long-input tasks beyond its physical window. Our core contribution is a full post-training recipe that unifies data synthesis, training methodologies, and agent architectures. First, to address the scarcity of high-quality, complex long-context reasoning data, we developed a novel synthesis pipeline. Crucially, our approach moves beyond generating simple "needle-in-a-haystack" retrieval~\citep{niah,ruler} or single-hop RAG~\citep{kovcisky2018narrativeqa} tasks, focusing instead on creating challenges that require \emph{multi-hop grounding and reasoning over globally distributed evidence}. The underlying principle is to deconstruct source documents into atomic facts and their relationships, and then programmatically compose complex, verifiable questions from this structured information. This scalable and principled synthesis strategy forms a critical foundation for \modelname, yielding substantial performance gains over the baseline, particularly on tasks demanding information aggregation and multi-hop reasoning. 
Second, we introduce several RL strategies to tackle the instability inherent in long-context, multi-task training. Validated through extensive in-depth experiments, our approach proposes two key improvements: (1) To counteract unstable mini-batch data distributions and mitigate reward estimation bias during multi-task training, we implement a \emph{task-balanced sampling} and \emph{task-specific advantage estimation} strategy. (2) To manage the critical exploration-exploitation trade-off, we propose the \emph{Adaptive Entropy-Controlled Policy Optimization} (AEPO) algorithm. It employs an entropy-based mechanism to actively control negative gradients, enabling the model to sustain training on sequences of progressively increasing length, thereby ensuring the scalability of its reasoning abilities. Third, to address reasoning tasks that exceed the model's physical context window, we introduce a \emph{memory management framework} extending its operational range. By employing  a multi-stage fusion RL training paradigm, we synergistically combines both the model's single-pass reasoning within its 256K context window, and an iterative memory updating mechanism to handle tasks extending beyond it.

Our comprehensive evaluation on leading long-context reasoning benchmarks~\citep{MRCR,longbench_v1,longbench_v2,krishna2024fact,zhao2024docmath} reveals substantial performance gains. As shown in Figure~\ref{fig:overall_results} and Table~\ref{tab:main}, \modelname surpasses its Qwen3-30B-A3B-Thinking-2507 baseline by an average of 9.9 points, achieving performance comparable to top-tier flagship models like GPT-5 and Gemini-2.5-Pro. This superiority is most evident on tasks demanding multi-hop grouding and reasoning over globally distributed evidence, validating the effectiveness of our data synthesis pipeline. Furthermore, the proposed multi-stage RL training paradigm to fuse the ability of single-pass reasoning and memory management proves to be effective,  as \modelname, operating in its memory-agent framework, outperforms the baseline's single-pass configuration by 15.26 points and its memory-agent configuration by 9.48 points on 1M$\sim$4M token tasks.
Crucially, the benefits of our proposed methods extend beyond long-context benchmarks. We observe \modelname's improvements across several key general domains, including mathematics and natural science question answering, tool-using agents, and long dialogue scenarios. This finding suggests that enhancing a model's long-context reasoning capabilities provides a foundational boost to its ability to maintain coherence and reason over extended informational sequences, a benefit that transcends specific task formats.

\section{Preliminary}
\subsection{Long-Context Reinforcement Learning}

We formulate the long-context reasoning task as a reinforcement learning (RL) problem. Given a set of $n$ documents $\{c_i\}_{i=1}^n$ and a question $q$, the goal of long-context RL is to optimize a policy model $\pi_{\theta}$ to generate a response $y$ that maximizes a reward function $r_{\phi}(c,q,y)$. The standard objective is to maximize the KL-regularized expected reward~\citep{schulman2017equivalence,qwenlongl1}:
\begin{equation}
\label{eq:vanilla_rl}
\max_{\pi_\theta} \mathbb{E}_{c,q\sim \mathcal{D}, y \sim \pi_{\theta}(\cdot \mid c,q)} 
\left[ r_{\phi}(c,q,y) \right] 
- \beta \mathbb{D}_{\text{KL}} \left[ \pi_{\theta}(y \mid c,q) \,||\, \pi_{\text{ref}}(y \mid c,q) \right],
\end{equation}
where $c = \text{Concat}(c_1,c_2, \dots,c_n)$, $\mathcal{D}$ is the training dataset, $\pi_{\text{ref}}$ denotes a reference policy, and $\beta$ controls the strength of the KL regularization to prevent large deviations from the reference policy.
\vspace{-0.24cm}

\subsection{Group Relative Policy Optimization (GRPO)} 
For long-context inputs, the quadratic complexity of the attention mechanism renders PPO~\citep{schulman2017proximal}, which relies on generalized advantage estimation  (GAE)~\citep{schulman2015high} via a value network, computationally prohibitive. 
Therefore, we employ GRPO~\citep{shao2024deepseekmath} to optimize the objective in Eq.~(\ref{eq:vanilla_rl}). 
For each input $(c, q)$, GRPO first samples a group of $G$ candidate responses $\{y_i\}_{i=1}^G$ from the old policy $\pi_{\theta_{\text{old}}}$. It then estimates the advantage through group-wise reward \textit{z}-score normalization, thereby obviating the need for a separate value network. 
Formally, the objective is:
\begin{equation}
\label{eq:grpo_objective}
\begin{aligned}
\mathcal{J}_\text{GRPO}(\theta) & = \mathbb{E}_{c,q \sim \mathcal{D}, \{ y_i \}_{i=1}^{G} \sim \pi_{\theta_\text{old}}( \cdot| c,q)} \Bigg[ \frac{1}{G}\sum_{i=1}^{G} \frac{1}{|y_i|}\sum_{t=1}^{|y_i|} \Bigg( \min \Big(\rho_{i,t}(\theta) A_i, \\&
\text{clip} \Big( \rho_{i,t}(\theta), 1 - \varepsilon, 1 + \varepsilon \Big) A_i \Big) - \beta \mathbb{D}_{\text{KL}}(\pi_{\theta} || \pi_{\text{ref}}) \Bigg) \Bigg],
\end{aligned}
\end{equation}
where $\rho_{i,t}(\theta) = \frac{\pi_{\theta}(y_{i,t} | c,q, y_{i,<t})}{\pi_{\theta_{\text{old}}}(y_{i,t} | c,q, y_{i,<t})}$ is the importance sampling ratio for token $t$ in sequence $i$. The group-relative advantage $A_{i}$ is shared across tokens of the $i$-th sequence and computed by normalizing the sequence-level rewards $\{r_i\}_{i=1}^G$:
\begin{equation}
\label{eq:advantage}
A_i = \frac{r_i - \text{mean}(\{r_k\}_{k=1}^G)}{\text{std}(\{r_k\}_{k=1}^G)}.
\end{equation}

To enhance stability and practical performance, we integrate a key technique by Decoupled Clip and Dynamic Sampling Policy Optimization (DAPO)~\citep{yu2025dapo}.
Specifically, we adopt a \textbf{token-level policy gradient loss}, which normalizes each token's contribution by the total number of tokens in the group.
This approach ensures every token in the same group contributes equally to the final objective, which prevents the learning signal from valuable tokens in high-quality, long responses from being diluted while ensuring that undesirable patterns in low-quality, lengthy outputs are effectively penalized. 

Consistent with recent findings suggesting that removing KL regularization can improve exploration and accelerate convergence~\citep{OpenReasonerZero2025, yu2025dapo, qwenlongl1}, we set $\beta=0$. 
Besides, we operate in a strictly on-policy setting, performing only a single gradient update per batch of samples. 
This design choice implies that the policy being updated, $\pi_{\theta}$, remains identical to the policy that generated the data, $\pi_{\theta_{\text{old}}}$. Since the importance sampling ratio $\rho_{i,t}(\theta)$ is strictly equal to 1, the clipping function becomes inactive, and we can remove it from the objective. 
Note that the advantage $A_{i}$ is independent of $t$, the training objective in Eq.~(\ref{eq:grpo_objective}) simplifies to:
\begin{equation}
\label{eq:grpo_objective_new}
\mathcal{J}_\text{GRPO}(\theta) = \mathbb{E}_{c,q \sim \mathcal{D}, \{y_i\}_{i=1}^{G} \sim \pi_{\theta_\text{old}}} \left[ \frac{1}{\sum_{j=1}^{G}|y_j|}\sum_{i=1}^{G}A_{i}\sum_{t=1}^{|y_i|} \rho_{i,t}(\theta)  \right].
\end{equation}

\subsection{Memory Agent}\label{sec:pre_mem_agent}
\begin{figure}[!t]
    \centering
    \includegraphics[width=0.99\textwidth]{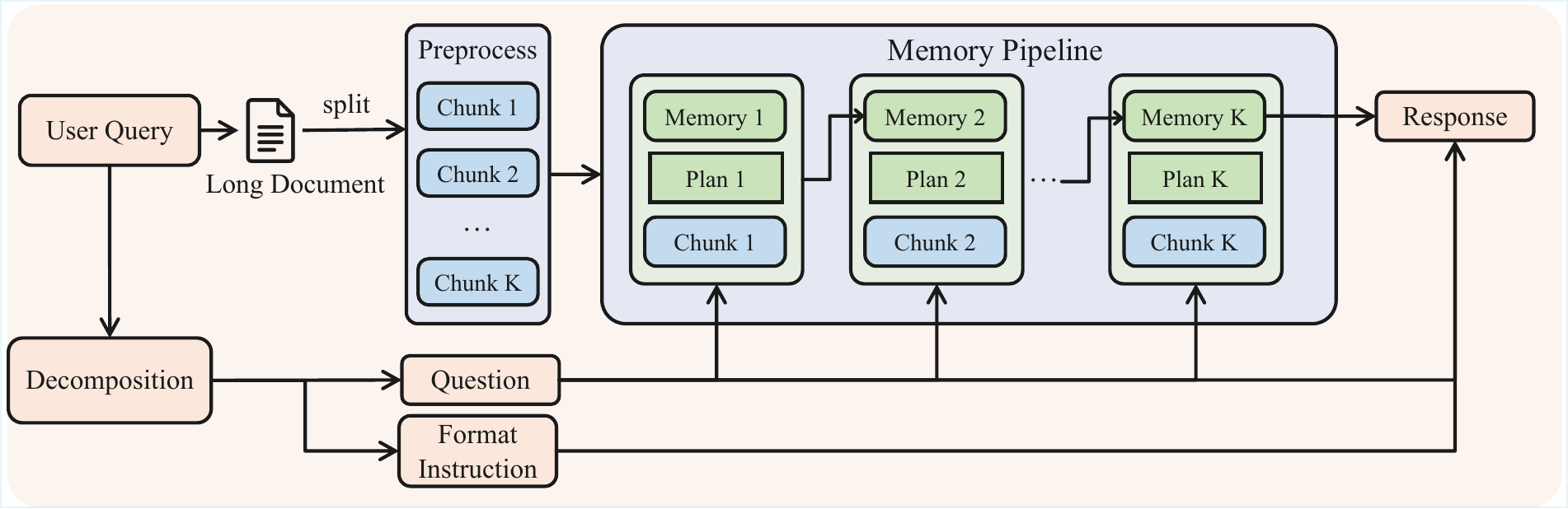}
    \vspace{-0.1cm}
    \caption{Memory agent workflow for processing ultra-long contexts}
    \label{fig:memory_workflow}
    \vspace{-0.25cm}
\end{figure}

To scale reasoning capabilities to ultra-long contexts where full-attention mechanisms become computationally intractable, we adopt the Memory Agent paradigm proposed by~\citet{yu2025memagent}. As illustrated in Figure~\ref{fig:memory_workflow}, this framework reframes the reading comprehension task as a sequential decision-making process rather than a single-pass inference.

\textbf{Sequential Memory Processing.}
Given a long context and a query, we first decompose the user query into a core question $q_{\text{core}}$ and formatting instructions $q_{\text{inst}}$. This separation prevents format constraints (e.g., JSON schema) from interfering with flexible reasoning during iterative memory updates. While $q_{\text{inst}}$ is reserved for the final generation to ensure format compliance, $q_{\text{core}}$ guides the reasoning process to maintain focus on answering the question.
The document is partitioned into chunks $\{x_1, \dots, x_K\}$. At each step $t$, the policy $\pi_{\theta}$ observes the current chunk $x_t$ and the historical states to update the memory $m_t$. We enhance the model by incorporating an explicit \textit{planning} mechanism: alongside the memory update, the agent generates a navigational plan $p_t$ to guide the attention and information extraction for the \textit{subsequent} chunk $x_{t+1}$. The state transition is formalized as:

\begin{equation}
\label{eq:mem_update}
(m_t, p_t) \sim \pi_{\theta}(\cdot \mid m_{t-1}, p_{t-1}, x_t, q_{\text{core}}).
\end{equation}

This recurrent mechanism effectively "folds" the global context into a compact representation while actively planning the reasoning path.

\textbf{Memory Optimization via RL.}
After processing the final chunk $x_K$, the model generates the final answer $y$ by integrating the accumulated memory $m_K$ with the original formatting instructions:

\begin{equation}
\label{eq:mem_answer}
y \sim \pi_{\theta}(\cdot \mid m_K, q_{\text{core}}, q_{\text{inst}}).
\end{equation}


To optimize the policy $\pi_{\theta}$, we employ the GRPO strategy based on trajectory-level rewards. Specifically, for each question $(q_{\text{core}}, q_{\text{inst}})$ and document chunks $x = \{x_1, \dots, x_K\}$, we sample $G$ distinct trajectories $\{\tau_1, \dots, \tau_G\}$. Each trajectory follows the pattern: at step $t \leq K$, the agent processes chunk $x_t$ to produce memory update $m_{i,t}$ and navigational plan $p_{i,t}$ for the next chunk; and finally produces answer $y_i$. Thus:
$$\tau_i = \{(m_{i,1}, p_{i,1}), \dots, (m_{i,K}, p_{i,K}),  y_i\}$$
A trajectory-level reward $R(\tau_i)$ is computed based on the correctness of $y_i$. We calculate the trajectory-level advantage via Eq.~(\ref{eq:advantage}) and broadcast it as a consistent learning target to all actions in the trajectory. The policy is updated end-to-end according to Eq.~(\ref{eq:grpo_objective_new}).

\section{Long-Context Data Construction}\label{sec:data_construct}


\begin{table}[!t]
\centering
    \caption{Comparison of RL training data characteristics between QwenLong-L1 and QwenLong-L1.5.}
    \label{tab:data_comparison}
    \resizebox{0.85\textwidth}{!}{
        \begin{tabular}{l>{\raggedright\arraybackslash}p{5cm}>{\raggedright\arraybackslash}p{6cm}}
        \toprule
         & \textbf{QwenLong-L1} & \textbf{QwenLong-L1.5} \\
        \midrule
        \textbf{Data Source} & Open-source & Open-source, Synthetic \\
        \midrule
        \textbf{\# Training Samples} & 1.6K & 14.1K \\
        \midrule
        \textbf{Domain} & $\bullet$ Professional documents \newline $\bullet$ General knowledge & $\bullet$ Code repositories \newline $\bullet$ Academic literature \newline $\bullet$ Professional documents \newline $\bullet$ General knowledge and literature \newline $\bullet$ Dialogue data \\
        \midrule
        \textbf{Question Type} & $\bullet$ Multi-fact reasoning \newline $\bullet$ Numerical calculation & $\bullet$ Multi-fact reasoning \newline $\bullet$ Numerical calculation \newline $\bullet$ Hypothetical scenarios \newline $\bullet$ Long in-context learning \newline $\bullet$ Temporal reasoning \newline $\bullet$ Causal analysis \newline $\bullet$ Viewpoint analysis \newline $\bullet$ Dialogue NIAH \newline $\bullet$ ... \\
        \midrule
        \textbf{Max Input Length} & 59,563 tokens &  119,932 tokens  \\
        \midrule
        \textbf{Avg. Input Length} & 11,441 tokens & 34,231 tokens  \\
        \bottomrule
        \end{tabular}
    }
\end{table}

\begin{figure}[!t]
    \centering
    \includegraphics[width=0.6\textwidth]{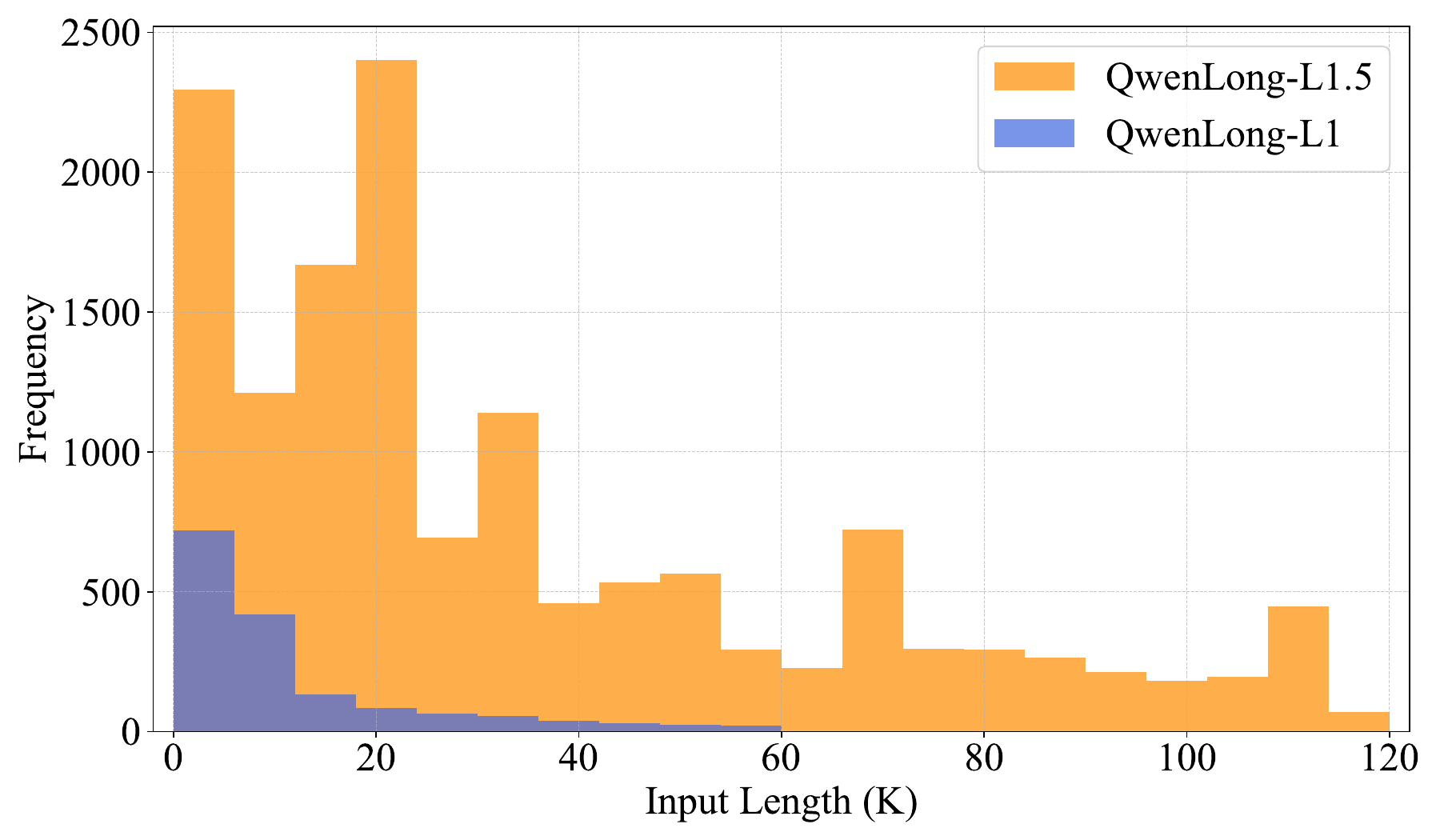}
    \vspace{-0.1cm}
    \caption{Comparison of Training Sample Input Length Distributions between QwenLong-L1 and QwenLong-L1.5.}
    \label{fig:token_distribution}
    \vspace{-0.25cm}
\end{figure}

For \modelname, we conducted a comprehensive scale-up of RL data tailored for long-context scenarios. Our efforts focused on enhancing the dataset's scale, diversity, and complexity. After a rigorous pipeline of multi-stage difficulty filtering, deduplication, and test set decontamination, we curated a final set of 14.1k high-quality training samples from an initial pool of 42.7k synthesized examples. As show in Table~\ref{tab:data_comparison}, this represents a significant expansion in scale compared to its predecessor, QwenLong-L1~\footnote{\url{https://huggingface.co/datasets/Tongyi-Zhiwen/DocQA-RL-1.6K}}. Furthermore, our dataset encompasses a broader spectrum of domains and question types, including multi-hop reasoning, numerical calculation, long in-context learning, temporal analysis, viewpoint analysis, and dialogue memory. As illustrated in Figure~\ref{fig:token_distribution}, we also pushed the boundaries of input length by substantially increasing the volume of training data longer than 64K tokens, thereby increased the overall complexity of the training data.

A key methodological decision for our RL data construction was to adopt a large-scale data synthesis pipeline. We observed that for complex long-context tasks, human annotators struggle to both formulate challenging questions requiring extensive reasoning and exhaustively verify answers within contexts exceeding 32k tokens. To overcome these limitations, we leveraged large language models to construct our RL dataset. As shown in Figure~\ref{fig:qa_synthesis}, our end-to-end pipeline proceeds as follows:

\textbf{Corpus Collection:} We assembled a vast and diverse repository of long documents, sourced from both web crawling and public open-source corpora, followed by a multi-stage quality filtering process to retain high-quality content.

\textbf{QA Synthesis:} We employed specialized methods to generate question-answer pairs. The primary goal here is to generate tasks with high  learning value by increasing their difficulty and ensuring the necessary information is dispersed throughout the long-context, rather than being easily located. To achieve this, we developed three techniques: synthesizing tasks from structured data (via knowledge graph or knowledge table) and refining simpler questions (via a multi-agent self-evolved framework). Subsequently, we further extend the context to our target length by strategically inserting irrelevant documents. This step significantly increases the long-context reasoning difficulty, forcing the model to identify and utilize sparsely located information within a much larger context.

\textbf{Data Verification:} Finally, all synthesized samples underwent two critical validation checks to ensure their quality and relevance to long-context reasoning: (1) \textbf{Knowledge Grounding Check:} We temporarily removed the source document and tested if the model could still answer the question. Samples that could be answered correctly (i.e., relying on the model's internal knowledge) were filtered out to ensure the data specifically tests contextual reasoning. (2) \textbf{Contextual Robustness Check:} We expanded the context with irrelevant documents and verified the model's answer. Any sample where the answer accuracy (pass@k) dropped to zero was discarded. This ensures the question and its answer are robust not brittle to context changes.

This structured and rigorous pipeline allows us to generate a large-scale, high-quality, and challenging dataset optimized for long-context RL training.

\subsection{Corpus Collection and Preprocessing}
The foundation of our data synthesis pipeline is a multi-source, heterogeneous corpus of long documents. We gathered materials from five primary categories to ensure broad coverage of topics and formats:

 \begin{itemize}
\item \textbf{Code repositories}: High-starred, high-quality open-source code repositories, primarily in Python.
\item \textbf{Academic literature}: Scholarly papers and textbooks from disciplines such as STEM, medicine, law, and social sciences, as well as recent AI research papers from arXiv.
\item \textbf{Professional documents}: Corporate annual reports, financial statements, product manuals, medical textbooks, and government publications.
\item \textbf{General knowledge and literature}: Classic novels, detective stories, and comprehensive Wikipedia pages.
\item \textbf{Dialogue data}:  A small collection of multi-turn dialogues simulated by large language models to cover conversational scenarios.
\end{itemize}

Following collection, all documents underwent meticulous rule-based and LLM-as-a-judge filtering to ensure high quality, factual accuracy, and coherence. This process yielded a final repository of 82,175 high-quality documents, totaling approximately 9.2 billion tokens. This curated corpus served as the rich raw material for our subsequent QA synthesis efforts.


\subsection{Question-Answer Synthesis}
We identify that a critical prerequisite for fostering advanced long-context reasoning in LLMs is the capability to capture dispersed contextual information and perform long-horizon reasoning on multi-hop and computationally intensive problems. Our question-answer synthesis pipeline is therefore designed to efficiently scale up the training data that targets and demands these capabilities. Specifically, our overarching strategy is to first mine salient local information and their interconnections from within long documents, and then leverage their relational links to construct more complex QA tasks. We designed three distinct methods to handle different types of information and relational structures, corresponding to three key problem categories:

\begin{figure}[!t]
    \centering
    \includegraphics[width=0.99\textwidth]{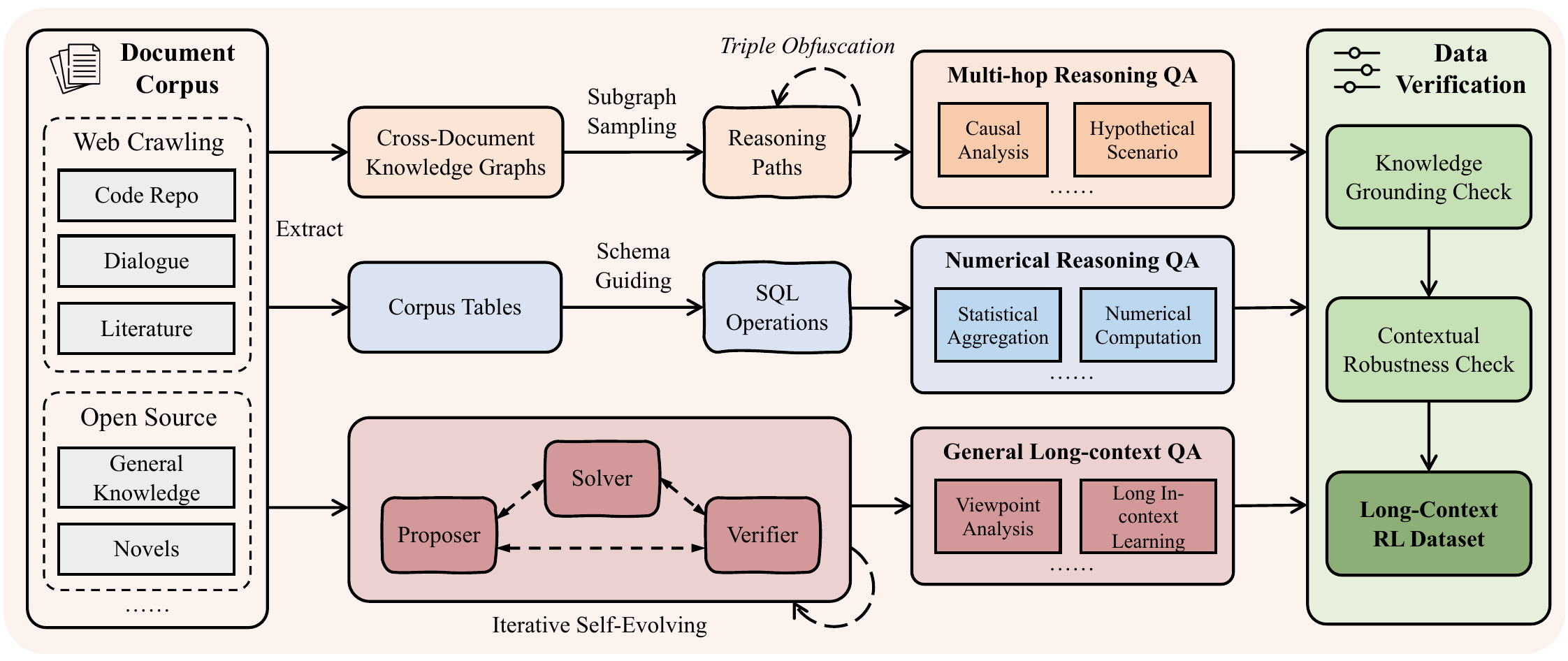}
    \vspace{-0.1cm}
    \caption{Overview of our end-to-end RL data synthesis framework}
    \label{fig:qa_synthesis}
    \vspace{-0.25cm}
\end{figure}

\paragraph{In-depth Multi-hop Reasoning QA:} To interconnect disparate information within documents and capture the long-range dependencies between them, we utilize the advantage of Knowledge Graph to mine complex reasoning paths and further construct challenging multi-hop questions based on them. Specifically, we employ a knowledge-graph-guided framework to synthesize in-depth reasoning QA through a three-stage process. First, KG Construction involves extracting triplets from documents across diverse domains to form an initial KG. This is expanded into a complex cross-document KG via domain-level aggregation and further refined using entity and relation clustering to ensure graph quality. Second, for Reasoning Path Sampling, we generate challenging multi-hop paths by sampling relation-relevant subgraphs centered on target entities. Long-range paths are derived using strategies such as Random Walk and BFS. To mandate rigorous cross-document information synthesis, path nodes are deliberately distributed sparsely across multiple documents. Furthermore, path complexity is heightened via information perturbation, including the obfuscation of entities (e.g., Temporal: "the year ending with 5 in the late 20th century" or Institutional: "a prestigious science university in Beijing"). Finally, in Question Generation, we synthesize multi-hop QA pairs based on the extracted paths, adopting a multi-paradigm approach that spans \textbf{Multi-fact Reasoning}, \textbf{Temporal Reasoning}, \textbf{Causal Analysis}, and \textbf{Hypothetical Scenarios}. We strictly control complexity by regulating path length and ensure quality through Blind Knowledge Screening and Scarce Knowledge Validation.

\paragraph{Corpus-level Numerical Reasoning QA: } The construction of high-quality, complex numerical reasoning questions across multiple, disparate documents often requires manual authoring and validation. To mitigate this reliance on manual effort, inspired by \citet{TongyiZhiwenTeam2024CorpusQA}, we introduce a Structural Tabular Data Engine designed to synthesize corpus-level numerical reasoning questions at scale. The process commences with Document Collection, where unstructured documents undergo parsing and subsequent filtering to ensure a prerequisite volume of tokens and the presence of rich statistical tables. Subsequently, we perform Schema Extraction to rigorously formalize the underlying data structure and relational schema. This formalized structure then enables Data Table Aggregation, transforming the disparate content into a unified, structured, cross-document corpus table. In parallel, a diverse pool of natural language Queries is generated through an LLM-based expansion from initial templates. These queries are then translated into executable SQL statements within the NL2SQL Execution stage. By executing the SQL against the aggregated tables, we accurately simulate complex calculation processes, such as statistical aggregation and numerical computation, thereby deriving the Ground Truth Answers. Finally, the relevant source documents are concatenated to form the Long Context QA Pair. These pairs are specifically designed to address complex scenarios involving \textbf{Statistical Aggregation}, \textbf{Numerical Calculation}, and \textbf{Temporal Reasoning}, which inherently necessitate advanced capabilities for global information integration and complex numerical reasoning.

\begin{figure}[!t]
    \centering
    \includegraphics[width=0.98\textwidth]{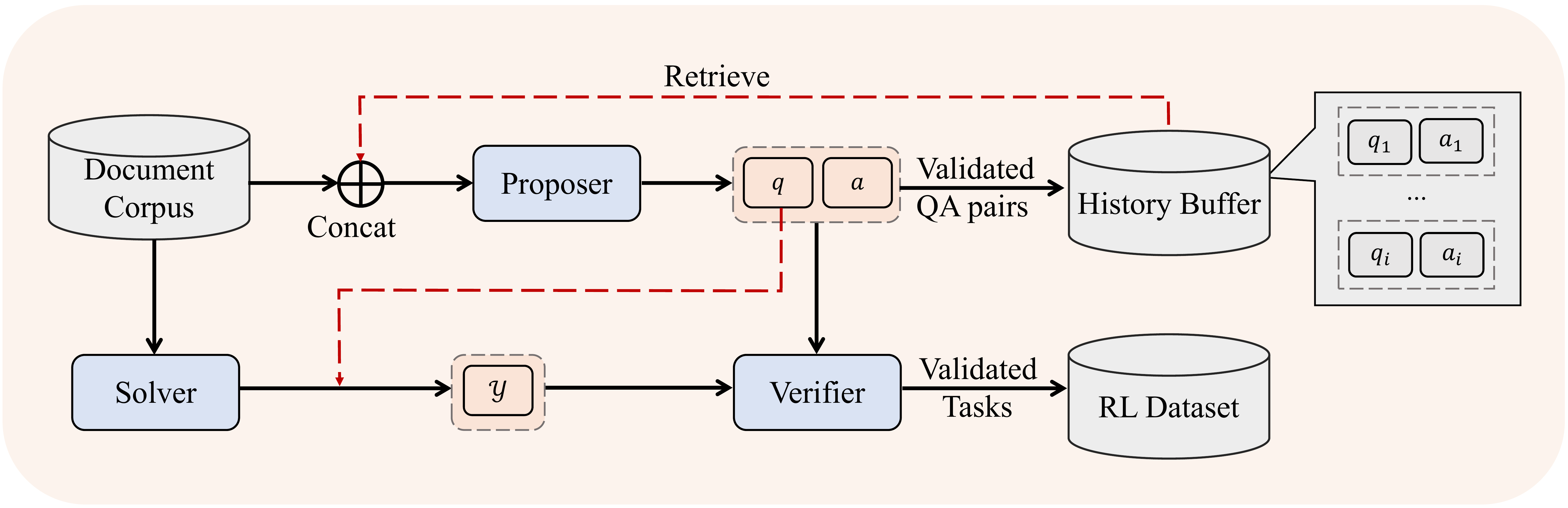}
    \vspace{-0.1cm}
    \caption{Overview of our proposed multi-agent self-evolved data synthesis framework.}
    \label{fig:mase}
    \vspace{-0.25cm}
\end{figure}

\paragraph{General Long-Context Reasoning: } For other general long-context tasks, such as \textbf{Viewpoint Analysis} and \textbf{Long In-context Learning}, we follow~\citet{SPELL} to introduce the multi-agent self-evolve (MASE) data synthesis framework, which automatically synthesizes long-context tasks and evolves their difficulty through the collaboration of three distinct agents. This method begins with proposing simple seed questions from the documents and iteratively increases their complexity and depth, covering a broad spectrum of general reasoning challenges.
As shown in Figure~\ref{fig:mase}, our MASE framework proceeds iteratively: given a cluster of $n$ documents $C=\{c_i\}_{i=1}^n$ and a task type $\tau$, a proposer agent first generates new questions, then a solver agent attempts to solve them, and finally a verifier agent validates the correctness of the generated QA pair. 

The proposer agent generates new question–answer pairs in an iterative curriculum. In the very first iteration, it is conditioned on all $n$ raw documents to produce a pair $(q, a)$. After each verified pair is created, we append it to a \textit{history buffer} $\mathcal{H}$ that stores the valid question–answer pairs corresponding to the current document set. In subsequent iterations, the proposer is conditioned on both the raw documents and the stored QA pairs.
These exemplars discourage redundancy and, via prompting, push the proposer to generate harder and more diverse questions than those already proposed. 
Then, the solver agent attempts to solve the generated question based on documents, and we employ a rule-based method to extract the final prediction $y$ from the response. 
Finally, the verifier agent estimates the semantic equivalence between the solver’s prediction $y$ and the proposer's reference answer $a$. Validated tasks are stored to the RL dataset, and corresponding QA pairs are saved to the history buffer.


\section{Long-Context Post-Training}\label{sec:rl_method}


\begin{figure}[!t]
    \centering
    \includegraphics[width=0.99\textwidth]{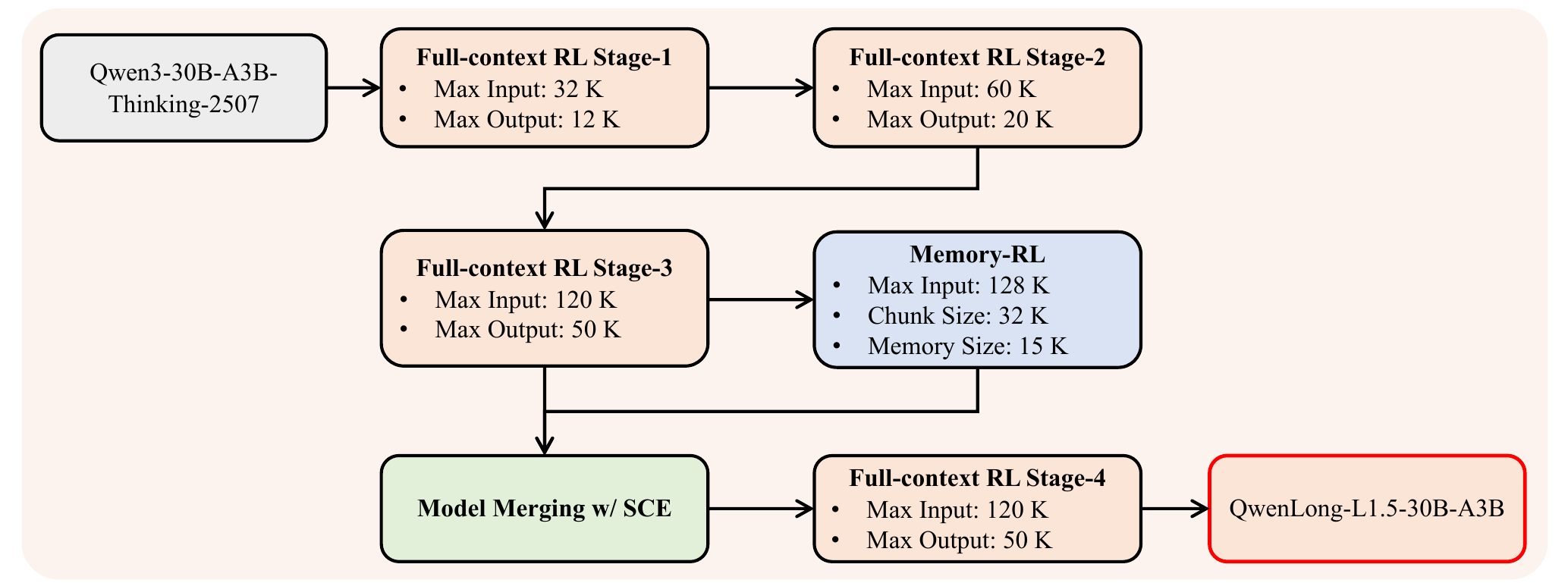}
    \vspace{-0.1cm}
    \caption{Post-training pipeline of QwenLong-L1.5.}
    \label{fig:training_paradigm}
    \vspace{-0.25cm}
\end{figure}

We introduce our overall post-training paradigm based on reinforcement learning for long-context reasoning in Section~\ref{sec:overall_paradigm}. In Sections~\ref{sec:multi_task_training} to~\ref{sec:AEPO}, we elaborate on the challenges encountered in long-context reasoning reinforcement learning, such as data-mix domain imbalance and training collapse, and present a series of optimization strategies for long-context reasoning RL along with the results of ablation experiments.

\subsection{Overall Progressive Training Paradigm}~\label{sec:overall_paradigm}

To avoid training instability caused by the abrupt transition from traditional short-input direct reasoning to the multi-hop grounding patterns required for long-context reasoning, we adopt the approach from QwenLong-L1~\citep{qwenlongl1}, employing a paradigm of multi-stage length extension to progressively enhance the model's long-context reasoning capabilities. Furthermore, we observe that as the input context length increases, the reasoning content length exhibits a generally positive growth trend. Therefore, to accommodate the output length requirements of tasks with varying input lengths, in \modelname, we synchronously extend the maximum rollout length for each RL stage. Specifically, we employ three different settings: (1) 20K tokens input with 12K tokens output; (2) 60K tokens input with 20K tokens output; and (3) 120K tokens input with 50K tokens output. During the transition between different RL stages, we adopt the difficulty-aware retrospective sampling strategy from QwenLong-L1, applying difficulty filtering to the training data using the input-output length settings of the subsequent stage.

As introduced in Section~\ref{sec:pre_mem_agent}, \modelname also integrates memory management capabilities. However, we found that mixing memory management training data and single-pass full-context processing training data together causes considerable damage to the overall RL training infrastructure efficiency and model training stability. Therefore, we adopted a paradigm of training specialized experts followed by model merging. Specifically, after 3 stages of full-context RL training, we continued memory management RL training based on QwenLong-L1.5-RL-Stage3 to obtain an expert model specialized in memory management, and utilized the SCE algorithm~\citep{fusechat} to merge this expert model with QwenLong-L1.5-RL-Stage3. Then, we performed full-context training on the merged model in the fourth stage, ultimately obtaining the QwenLong-L1.5 model. Our overall training pipeline is illustrated in Figure~\ref{fig:training_paradigm}. In Section~\ref{sec:exp_multi_stage}, we compare the performance of models at different stages and find that the model's long-context reasoning capability continuously evolves through multi-stage training, ultimately resulting in a model that possesses both long-context capabilities and memory management capabilities.

\subsection{Multi-Task Reinforcement Learning}\label{sec:multi_task_training} 

\begin{figure}[!t]
    \centering
    \includegraphics[width=0.85\textwidth]{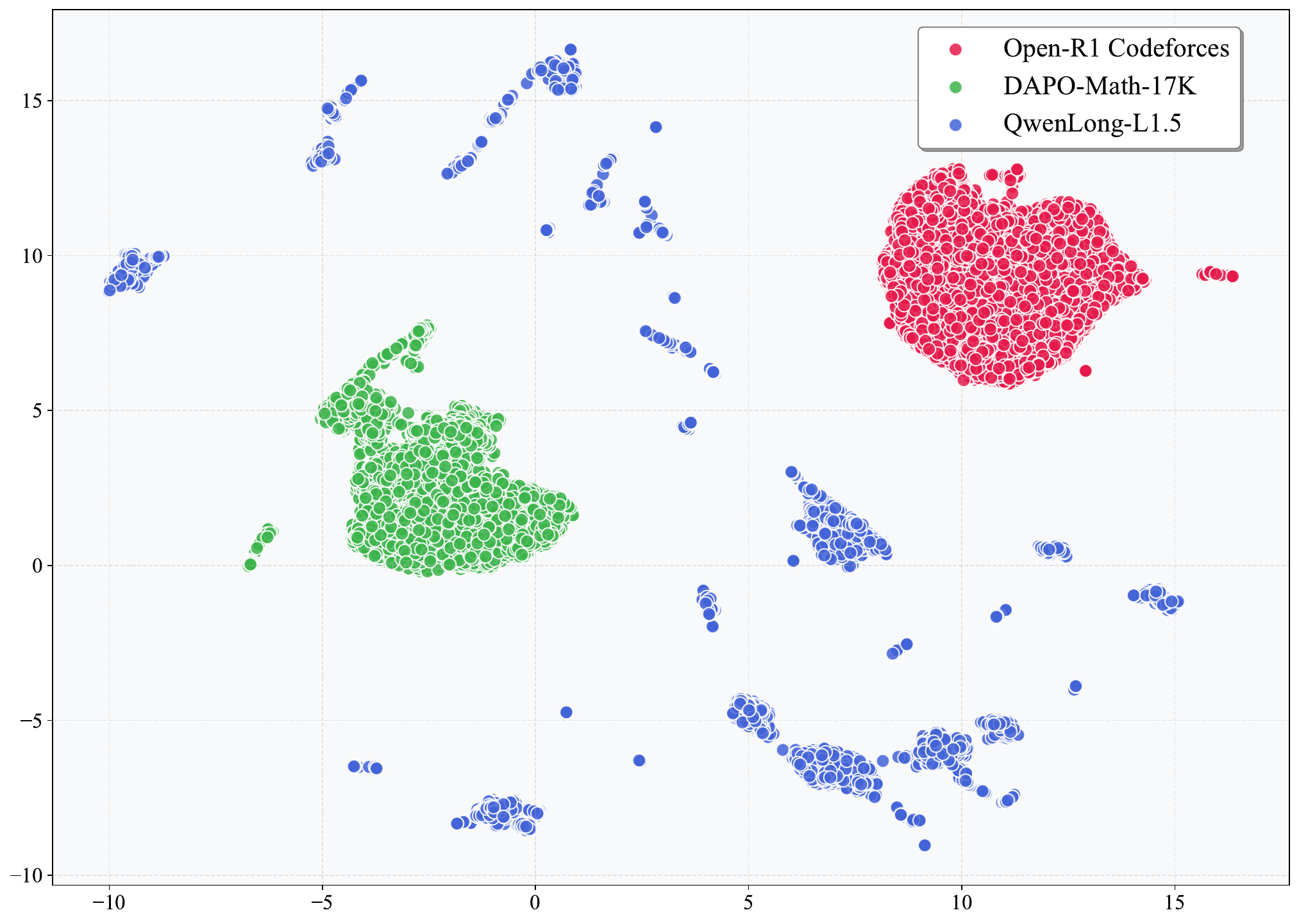}
    \vspace{-0.1cm}
    \caption{Two-dimensional UMAP projection~\citep{mcinnes2018umap} of three datasets  (Open-R1 Codeforces~\citep{openr1}, DAPO-Math-17K~\citep{yu2025dapo}, and QwenLong-L1.5 training set) using Qwen3-30B-A3B-Thinking-2507 embeddings.}
    \label{fig:data_vis}
    \vspace{-0.25cm}
\end{figure}

\begin{figure}[!t]
    \centering
    \includegraphics[width=0.99\textwidth]{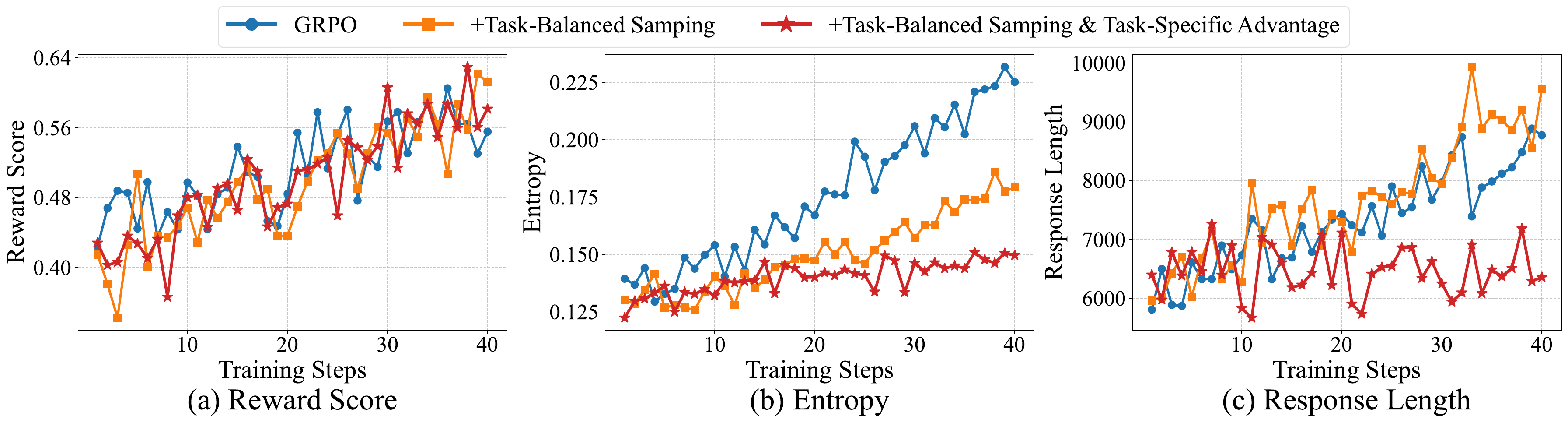}
    \vspace{-0.1cm}
     \caption{Comparison of training dynamics. By integrating task-balanced sampling with task-specific advantage estimation, our method achieves similar reward growth as the GRPO baseline while ensures a more stable training dynamics, evidenced by the stabilized entropy and the controlled response length growth.}
    \label{fig:taskbalanced}
    \vspace{-0.25cm}
\end{figure}

\paragraph{Task-balanced sampling} As described in Section~\ref{sec:data_construct}, long-context data, due to its diverse problem types and contextual domains, is more susceptible to distributional drift compared to traditional short-input data. As illustrated in Figure~\ref{fig:data_vis}, in contrast to traditional RL training data for reasoning tasks such as mathematics and code, long-context data exhibits a multi-clustered distribution, characterized by significant divergence among the different clusters. Based on this characteristic, a natural implication for RL training is that a traditional random sampler can lead to distributional imbalances within each training batch, thereby compromising training stability. This is specifically manifested as a rapid increase in the instantaneous entropy of the baseline (as shown in Figure~\ref{fig:taskbalanced}), which in turn prevents the training process from scaling up effectively. To this end, we implemented several strategies within our RL framework to ensure the balance of training samples:

\begin{itemize}
\item Prior to training, we perform balanced sampling on training data from different domains and task types. We first perform pre-inference on the data from each source using the base model. We then stratify the data into uniform bins based on the resulting pass@k scores. Finally, an equal number of training samples are uniformly sampled from each bin.
\item During training, we replace the conventional random sampler in the RL framework with a task-balanced sampler. In the sampling process for each training batch, this sampler draws an equal number of samples from each of the five designated task types: multiple choices, doc multi-hop reasoning, general reading comprehension, dialogue memory, and corpus-level numerical calculation.
\end{itemize}

\paragraph{Task-specific advantage estimation} In GRPO, the group-level reward standard deviation normalization used for advantage estimation can introduce a biased estimation~\citep{liu2025understanding}. Consequently, batch-level normalization was proposed to mitigate this bias~\citep{hu2025reinforce++,liu2025part}. However, given our task-balanced sampling strategy, this batch-level normalization can, in turn, introduce potential noise into the advantage estimation due to the reward distribution variance
across different tasks. 
Therefore, we adopt a task-aware approach to compute the reward standard deviation when estimating advantage. Specifically, for the $i$-th response of the policy model, we modify the group-level standard deviation in Eq.(\ref{eq:advantage}) to the standard deviation of rewards from all samples belonging to the same task within the current training batch $\mathcal{B}^{\text{task}}$:
\begin{equation}
A_{i}^{\text{task}} = \frac{r_{i}^{\text{task}} - \operatorname{mean}(\{r_{k}^{\text{task}}\}_{k=1}^{G})}{\textcolor{red}{\operatorname{std}(r^{\text{task}} | r^{\text{task}} \in \mathcal{B}^{\text{task}})}}, \quad \text{task} \in \{\text{mc, qa, niah, \dots}\}
\end{equation}

Compared to the group-level approach, this task-level method reduces the bias caused by noisy samples. 
In contrast to the batch-level standard error estimation, it isolates tasks with dense reward (e.g., NIAH tasks with reward range from 0 to 1) and sparse reward (e.g., qa or multiple-choice tasks with reward range within 0 and 1), thus providing a more accurate estimation for different tasks.

\begin{table}[!t]
\centering
    \caption{Ablation experiments of multi-task reinforcement learning strategies.}
    \label{tab:multi_task_rl}
        \resizebox{0.999\textwidth}{!}{
        \begin{tabular}{lccccccc}
        \toprule
        \textbf{Models}  & \textbf{Avg.} & \textbf{DocMath} & \textbf{LBV2} & \textbf{Frames}  & \textbf{MRCR} & \textbf{CorpusQA} & \textbf{LBV1-QA} \\
        \midrule
        Qwen3-4B-Thinking-2507 & 52.79&59.00&41.35&62.86&39.85&49.38&64.30  \\
        \ \ + GRPO & 56.07& 61.25&44.33&67.11&40.90&58.75&64.10 \\
        \midrule
        \rowcolor{Gray}\ \ + Task-balanced sampling &56.86&59.38&42.94&66.87&51.66&56.25&64.10 \\
        \rowcolor{Gray}\ \ + Task-balanced sampling + Batch-std &57.45&62.00&45.53&65.78&48.43&60.31&62.70 \\
        \rowcolor{Gray}\ \ + Task-balanced sampling + Task-batch-std & \textbf{58.62}&61.25&43.54&67.48&53.23&60.94&65.30\\
 
        \bottomrule
    \end{tabular}
    }
\end{table}

\paragraph{Experiments}  We conduct experiments on Qwen3-4B-Thinking to compare our two proposed enhancements with traditional GRPO.
Figure~\ref{fig:taskbalanced} illustrates the training dynamics, and Table~\ref{tab:multi_task_rl} presents the evaluation results across multiple long-context benchmarks.
Integrating task-balanced sampling with GRPO leads to more stable entropy dynamics compared to the baseline. However, this combination still exhibits significant growth in response length, resulting in only a moderate overall performance improvement.
When task-balanced sampling is further combined with task-specific advantage estimation, the training process becomes more stable. This combination achieves reward growth comparable to the GRPO baseline while notably stabilizing entropy and effectively controlling the increase in response length. We attribute this improvement to two main factors: (1) the task-specific standard deviation estimation generally yields a larger normalization denominator, which prevents excessive gradient update. (2) calculating reward variance separately per task-batch accommodates the distinct reward distributions across different tasks, providing a more accurate and isolated advantage estimation for tasks with dense reward and sparse reward.
This combined approach achieves an average of 2.55-point gain compared to the GRPO baseline. Notably, the improvement is particularly pronounced on MRCR, which is a task with dense reward.

\subsection{Negative Gradient Clipping} 
\label{sec:token-level-ngc}

\begin{figure}[h]
    \centering
    \begin{minipage}[b]{0.25\textwidth}
        \centering
        \includegraphics[width=\textwidth]{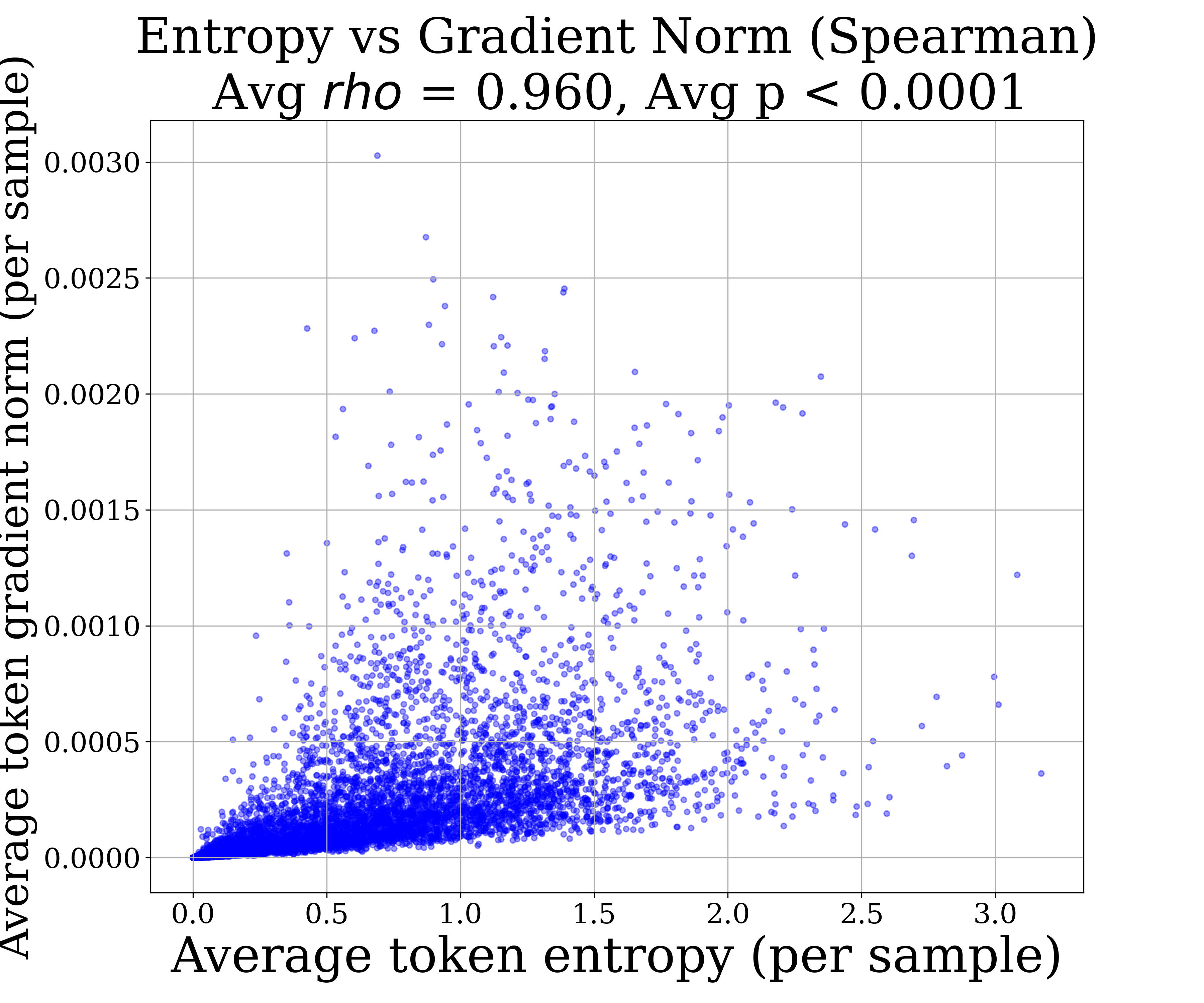}
        \caption{Correlation between token entropy and gradient norm in negative rollouts. The Spearman’s $\rho=0.96 (p<0.0001)$}
        \label{fig:entropy_vs_grad}
    \end{minipage}
    \hfill
    \begin{minipage}[b]{0.72\textwidth}
        \begin{subfigure}[b]{0.48\textwidth}
            \centering
            \includegraphics[width=\textwidth]{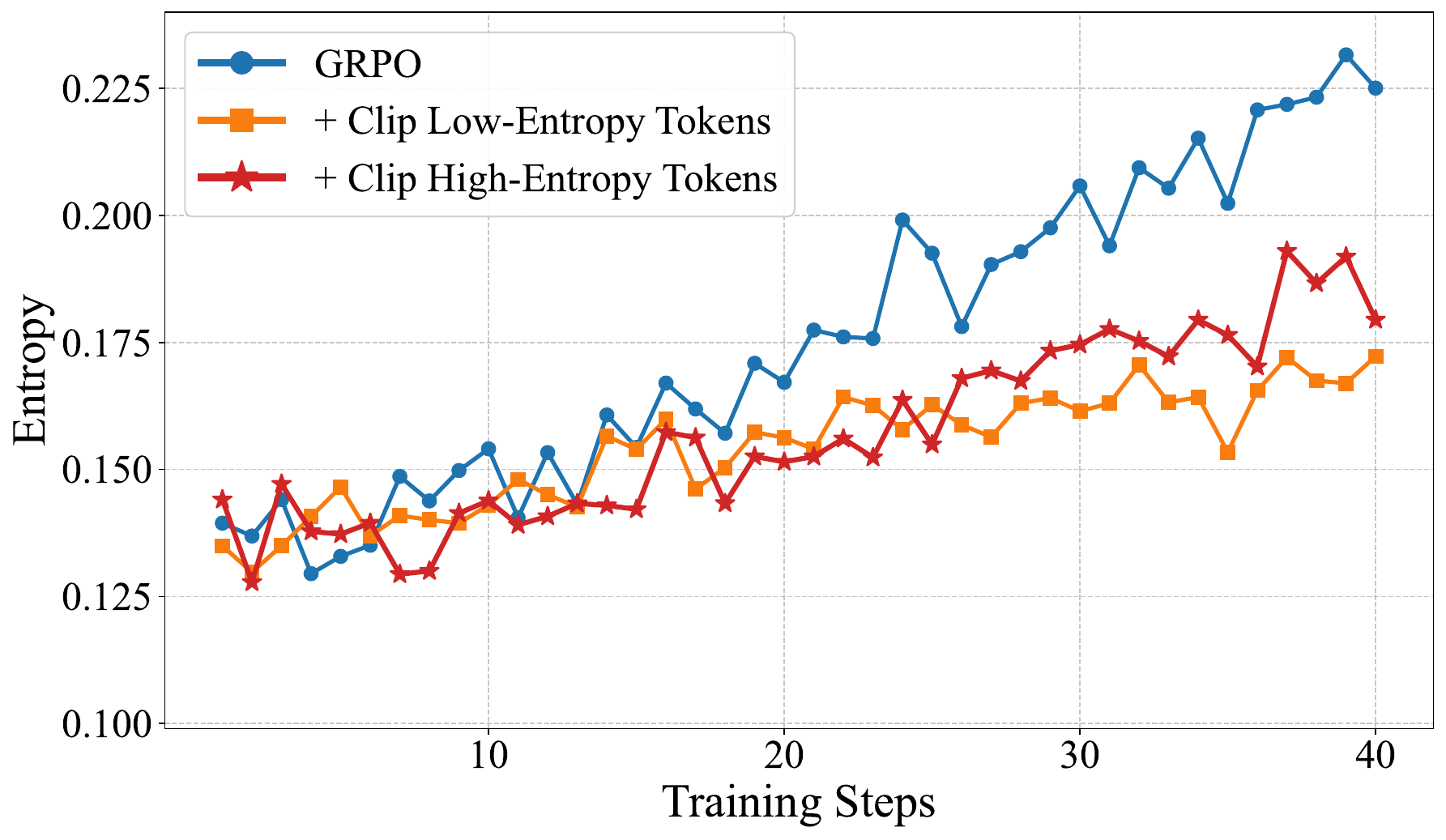}
            \caption{Token-level. Clipping either low- or high-entropy tokens reduces the model’s overall entropy, but the results in Table~\ref{tab:performance_negative_clipping} shows that clipping high-entropy tokens yields larger gains.}
            \label{fig:token_clipping_dynamic}
        \end{subfigure}
        \hfill
        \begin{subfigure}[b]{0.48\textwidth}
            \centering
            \includegraphics[width=\textwidth]{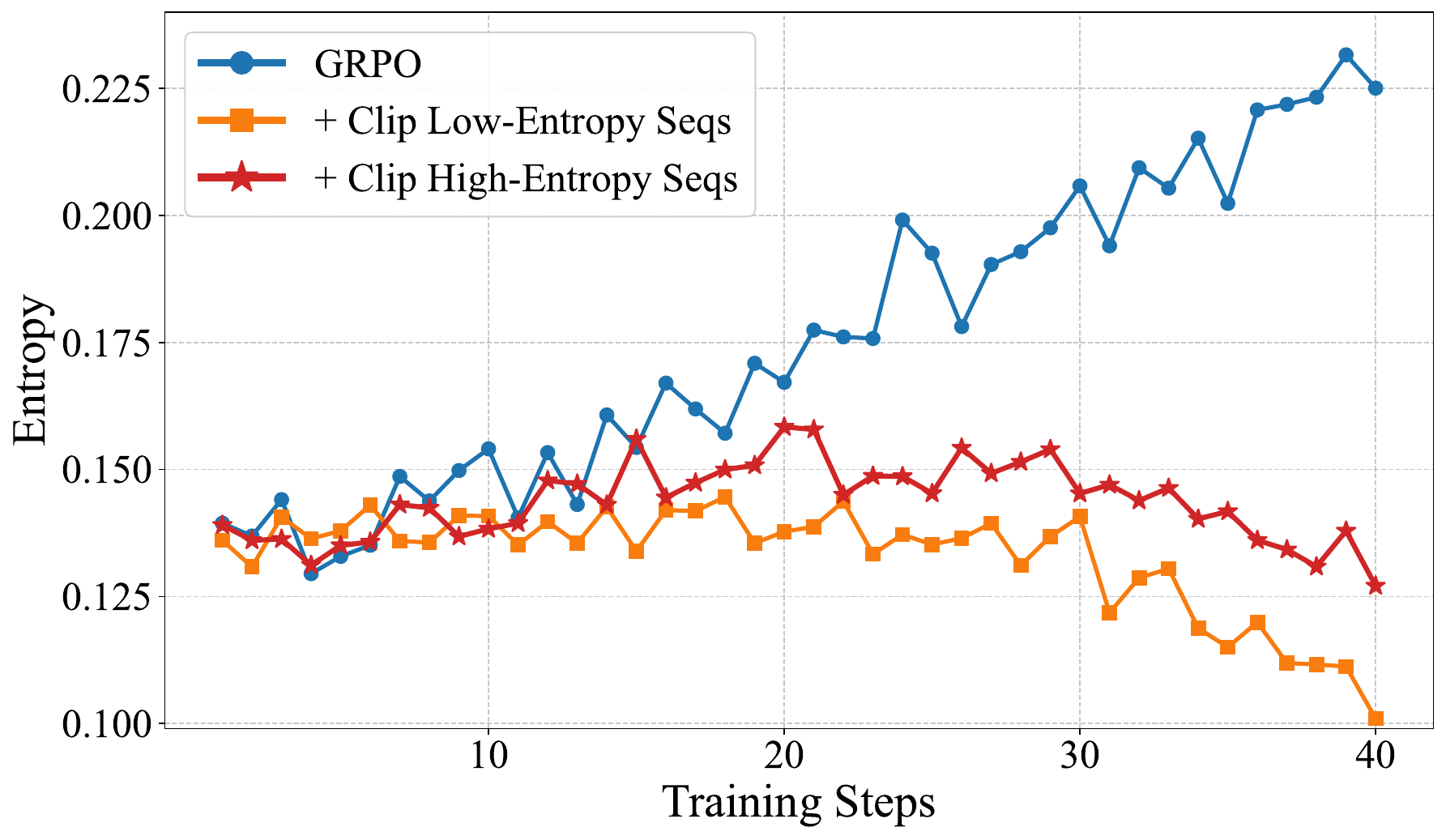}
            \caption{Sequence-level. Clipping either low- and high-entropy sequences reduces model entropy, and entropy decreases faster than with token-level clipping because more negative gradients are removed.}
            \label{fig:seq_clipping_dynamic}
        \end{subfigure}
        \caption{Entropy dynamics under different negative gradient clipping strategies.}
        \label{fig:clipping_dynamic}
    \end{minipage}
\end{figure}

\paragraph{Motivation and approach}
Unlike short-context tasks (e.g., mathematics), long-context tasks require models first to locate task-relevant information within the context and then perform reasoning~\citep{fundamental_capability,solopo}. This constraint forces all responses to be grounded in the same context, resulting in a higher similarity between correct and incorrect reasoning paths. As shown in Table~\ref{tab:rouge_compa}, the phrase-level overlap between the model's correct and incorrect responses, quantified by ROUGE-L, is substantially greater for DocMath than for AIME24/25. This high similarity implies that incorrect responses contain numerous correct steps (see cases in Figure~\ref{fig:case_correct_steps_in_nega_steps} and Figure~\ref{fig:case_hig_entropy_2}), further exacerbating the reward credit assignment problem~\citep{Sutton1998ReinforcementLA,Arumugam2021AnIP} in RL, ultimately leading to training instability. To mitigate this issue, we attempt to clip a portion of the negative gradients generated by negative responses. Firstly, as illustrated in Figure~\ref{fig:entropy_vs_grad}, during long-context RL, a strong correlation exists between high-entropy tokens and their corresponding gradient norms. This phenomenon indicates that high-entropy tokens tend to produce large gradients, which can increase the variance of parameter updates and destabilize optimization. Moreover, high-entropy tokens often reflect exploratory behavior in the reasoning process~\citep{wang2025beyond}, and avoiding excessive penalization helps preserve the model’s ability to explore and potentially correct originally erroneous paths. Building upon these considerations, we propose to clip either high-entropy negative responses or high-entropy tokens within negative responses to stabilize the training process:


\begin{align}
\label{eq:grpo_objective_new_with_clipping}
\mathcal{J}_\text{GRPO}(\theta) = \mathbb{E}_{c,q \sim \mathcal{D}, \{y_i\}_{i=1}^{G} \sim \pi_{\theta_\text{old}}} \left[ \frac{1}{\sum_{j=1}^{G}|y_j|}\sum_{i=1}^{G}A_{i}\sum_{t=1}^{|y_i|} \rho_{i,t}(\theta)\mathbb{I}(t,i)  \right],
\end{align}
where $\mathbb{I}(i,t)$ is an indicator function, defined as:
\begin{align}
\label{eq:indicator_function}
\mathbb{I}(i, t) &= \begin{cases}
    0 & \text{if } A_i < 0 \text{ and } \left( (P_{\text{token\_level}} \land H(t|i) > \tau_{\text{token}}) \right. \\
      & \quad \quad \quad \quad \quad \quad \left. \lor (\neg P_{\text{token\_level}} \land \bar{H}(i) > \tau_{\text{sequence}}) \right) \\
    1 & \text{otherwise}
\end{cases}\\
\end{align}
Here, $H(t|i)=-\sum_{v\in V}\pi_\theta(y_{i,t}|c,q,y_{i,<t})\log\pi_\theta(y_{i,t}|c,q,y_{i,<t})$ and $\bar{H}(i)=\frac1{|y_i|}\sum_{t=1}^{|y_i|}H(t|i)$ denote the token-level and sequence-level entropies for response $i$ (at position $t$), respectively. $\tau_{\text{token}}$ and $\tau_{\text{sequence}}$ are their corresponding thresholds. $P_{\text{token\_level}}$ is a boolean parameter selecting between token-level (True) and sequence-level (False) clipping.

\begin{table}[h]
        \centering
        \caption{Similarity between correct and incorrect responses in short- and long-context settings.}
        \begin{tabular}{lcc}
        \toprule
            \textbf{Tasks} & \textbf{Avg. Input length} & \textbf{ROUGE-L} \\
        \midrule
            AIME24/25 &0.18K tokens & 27.71\\
            DocMath &20.1K tokens& 45.37\\
        \bottomrule
        \end{tabular}
        \label{tab:rouge_compa}
\end{table}

\paragraph{Experiments} 
We conduct experiments on Qwen3-4B-Thinking to compare the impact of different clipping strategies on training stability and performance.
Figure~\ref{fig:clipping_dynamic} illustrates the entropy dynamics during training, and Table~\ref{tab:performance_negative_clipping} reports the evaluation results across multiple long-context benchmarks. In the token-level setting, clipping either high- or low-entropy tokens helps stabilize training, but clipping low-entropy tokens yields lower final performance, indicating the importance of penalizing high-confidence actions in negative trajectories~\citet{wang2025harnessinguncertaintyentropymodulatedpolicy}. In particular, on MRCR, clipping low-entropy tokens results in a 10-point drop compared to clipping high-entropy tokens. In the sequence-level approach, both clipping strategies stabilize training and may boost performance. However, removing too many negative gradient signals can cause entropy collapse, reduce exploration, and ultimately degrade results. For instance, performance dropped between steps 30 and 40 for low-entropy sequence clipping.

Considering that, with a proper balance between exploration and exploitation, sequence-level clipping can better stabilize model optimization and provide additional performance gains, we further propose an iterative training paradigm based on sequence-level negative gradient clipping in Section~\ref{sec:AEPO}.

\begin{table}[h]
    \centering
    \caption{Ablation experiments of negative gradient clipping strategies}
    \resizebox{0.999\textwidth}{!}{
    \begin{tabular}{lccccccc}
        \toprule
        \textbf{Models}&\textbf{Avg.}&\textbf{DocMath}&\textbf{LBV2}&\textbf{Frames}&\textbf{MRCR}&\textbf{CorpusQA}&\textbf{LBV1-QA}\\
             \midrule
             Qwen3-4B-Thinking-2507&52.79&59.00&41.35&62.86&39.85&49.38&64.30\\
             \ \ + GRPO&56.07&61.25&44.33&67.11&40.90&58.75&64.10\\
             \midrule
             \multicolumn{8}{c}{\textbf{Token-Level Negative gradient Clipping}}\\
             \midrule
             \rowcolor{Gray}\ \ + clip \textbf{low} entropy \textbf{tokens}&55.56&62.12&44.73&66.50&36.29&59.53&64.20\\
             \rowcolor{Gray}\ \ + clip \textbf{high} entropy \textbf{tokens}&\underline{57.02}&62.12&44.14&67.35&46.20&59.69&62.60\\
             
             \midrule
             \multicolumn{8}{c}{\textbf{Sequence-Level Negative gradient Clipping}}\\
             \midrule
             
             \rowcolor{Gray}\ \ + clip \textbf{low} entropy \textbf{seqs(step 30)}&56.66&62.00&44.40&66.26&45.64&56.88&64.80\\
             \rowcolor{Gray}\ \ + clip \textbf{low} entropy \textbf{seqs(step 40)}&55.47&62.25&41.64&65.78&45.54&55.62&62.00\\
             \rowcolor{Gray}\ \ + clip \textbf{high} entropy \textbf{seqs}&\textbf{57.36}&62.38&45.18&68.69&41.31&61.88&64.70\\

             
             \bottomrule
        \end{tabular}
    }
    \label{tab:performance_negative_clipping}
\end{table}

\subsection{Adaptive Entropy-Controlled Policy Optimization (AEPO)}
\label{sec:AEPO}

Building upon the findings in Section~\ref{sec:token-level-ngc}, which identify negative advantages coupled with high entropy as the primary source of instability in long-context RL, we propose the Adaptive Entropy-controlled Policy Optimization (AEPO) algorithm. 
AEPO dynamically masks rollout sequences associated with negative advantages during online RL training. This masking is governed by the policy entropy, which quantifies the inherent randomness of the tokens generated by the policy model. Given a policy model $\pi_\theta$ and a training batch $\mathcal{B}$, the batch-level entropy is defined as:
\begin{equation}
    H(\pi_\theta, \mathcal{B}) = - \frac{1}{|\mathcal{B}|} \sum_{i=1}^{|\mathcal{B}|} \frac{1}{|y_i|} \sum_{t=1}^{|y_i|} \sum_{v\in V}\pi_\theta(v|c,q,y_{i,<t})\log\pi_\theta(v|c,q,y_{i,<t}),
\end{equation}
where $V$ denotes the vocabulary space of the policy model.
We establish a target entropy range for the policy, defined by a lower bound $H_{\text{low}}$ and an upper bound $H_{\text{high}}$. During training, if the average batch entropy exceeds $H_{\text{high}}$, AEPO masks all samples with negative advantages. The model is updated exclusively using positive samples, which is functioned as an advantage-weighted online rejection sampling fine-tuning, effectively reducing the model's entropy. Conversely, when the entropy drops below $H_{\text{low}}$, the negative gradients are reintroduced to the optimization process to prevent entropy collapse. 

\begin{figure}[!t]
    \centering
    \includegraphics[width=0.75\textwidth]{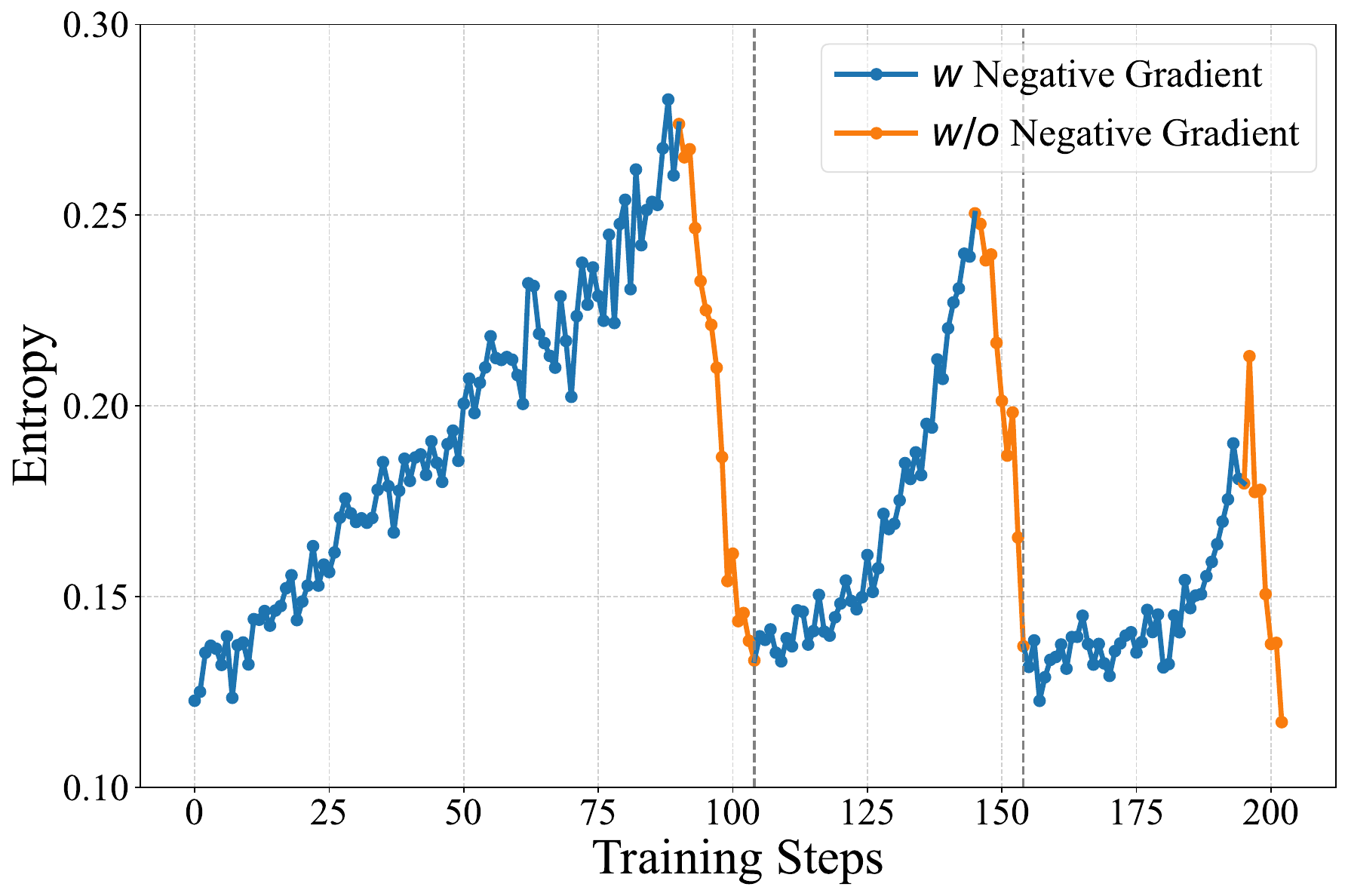}
    \vspace{-0.1cm}
    \caption{Entropy dynamics of AEPO algorithm on Qwen3-30B-A3B-Thinking.}
    \label{fig:entropy_dynamics}
    \vspace{-0.25cm}
\end{figure}

Through this dynamic entropy control, AEPO achieves significant improvements. As shown in Table~\ref{tab:aepo}, the method yields an average performance gain of 3.29 points over the GRPO baseline on Qwen3-4B-Thinking-2507. As illustrated in Figure~\ref{fig:entropy_dynamics}, when applied to our primary model, Qwen3-30B-A3B-Thinking, AEPO maintains an optimal balance between exploration (with negative gradient) and exploitation (without negative gradient). 
This stability is crucial for scaling RL training to a larger number of steps without degradation.

\begin{table}[h]
\centering
    \caption{Abalation experiments of AEPO algorithm on Qwen3-4B-Thinking-2507.}
    \label{tab:aepo}
        \resizebox{0.999\textwidth}{!}{
        \begin{tabular}{lccccccc}
        \toprule
        \textbf{Models}  & \textbf{Avg.} & \textbf{DocMath} & \textbf{LBV2} & \textbf{Frames}  & \textbf{MRCR} & \textbf{CorpusQA} & \textbf{LBV1-QA} \\
        \midrule
        Qwen3-4B-Thinking-2507 & 52.79&59.00&41.35&62.86&39.85&49.38&64.30  \\
        \ \ + GRPO & 56.07& 61.25&44.33&67.11&40.90&58.75&64.10 \\
        \midrule
        \rowcolor{Gray}\ \ + AEPO &\textbf{59.36}&62.50&47.91&67.35&47.88&64.69&65.80 \\

        \bottomrule
    \end{tabular}
    }
\end{table}

\section{Experiments}

\subsection{Experimental Setup}

\paragraph{Training Details}  
We select Qwen3-30B-A3B-Thinking-2507~\citep{qwen3_technical_report} as our base model for multi-stage RL training. 
Our RL framework is implemented using VeRL~\citep{verl}. During generation, we employ a sampling temperature of 0.7 and a top-$p$ value of 0.95. 
To balance rollout diversity and computational efficiency, we utilize a group size of $G=8$. 
We conduct a purely on-policy RL training with a batch size of 128 and a constant learning rate of $2 \times 10^{-6}$. 
Following QwenLong-L1~\citep{qwenlongl1}, we adopt a hybrid reward mechanism that combines rule-based verification with LLM-as-a-judge. Specifically, rule-based verification is used to check whether the ground truth is contained within the model output, while gpt-oss-120b~\citep{gpt_oss_system_card} is employed as an external judge for questions where rule-based verification is failed.


\paragraph{Evaluation Benchmarks} We evaluate our models using a suite of well-established benchmarks designed to assess long-context comprehension and reasoning. These benchmarks fall into three primary categories: multiple-choice, needle in a haystack (NIAH), and multi-hop question answering (QA).
For the multiple-choice task, we use LongBench-V2~\citep{longbench_v2}, a benchmark of 503 questions that assesses deep comprehension across six areas: 
single-document QA, multi-document QA, long in-context learning, long-dialogue history understanding, code repository understanding, and long structured data understanding. 
For NIAH, we use MRCR~\citep{MRCR}\footnote{\url{https://huggingface.co/datasets/openai/mrcr}}, which tests the model’s ability to find and disambiguate between multiple needles hidden in multi-turn synthetic conversations.
For multi-hop QA, our evaluation incorporates several benchmarks: Frames~\citep{krishna2024fact}, containing 824 questions on diverse Wikipedia topics such as history, sports, science, animals, and health; five subsets from LongBench~\citep{longbench_v1}, namely 2WikiMultihopQA~\citep{ho2020constructing}, HotpotQA~\citep{yang2018hotpotqa}, Musique~\citep{trivedi2022musique}, NarrativeQA~\citep{kovcisky2018narrativeqa}, Qasper~\citep{dasigi2021dataset}; DocMath~\citep{zhao2024docmath}\footnote{For DocMath, we use the test-mini subset of 800 queries.}, which focuses on numerical reasoning within financial reports; and CorpusQA~\citep{TongyiZhiwenTeam2024CorpusQA}, which challenges models to perform global reasoning over evidence dispersed throughout a large corpus of documents.

\vspace{-7pt}

\paragraph{Evaluation Configurations} We evaluate all models with
maximum input lengths of 128K tokens and maximum generation length of 50K tokens. 
For prompts exceeding the maximum context window, we employ the middle truncation strategy from ~\citet{longbench_v1} to preserve the front and tail portions of the context. 
All experiments are conducted using a sampling temperature of 0.7 and a top-$p$ value of 0.95. 
Our scoring is tailored to each benchmark's format. For multiple-choice tasks, we report standard accuracy. For MRCR, we report the SequenceMatcher ratio\footnote{\url{https://docs.python.org/3/library/difflib.html}}.
For multi-hop QA tasks, we report the maximum of cover exact match (CEM)~\citep{r1seacherplusplus} score and LLM-as-a-judge~\citep{zheng2023judging}, which uses DeepSeek-V3~\citep{liu2024deepseekv3} to evaluate semantic equivalence between a model's prediction and the ground-truth answer. 
The prompt for this evaluation is detailed in Table~\ref{tab:llm_judge}. 

\begin{table}[!ht]
    \centering
\caption{Prompt template for LLM-as-a-judge to compare the semantic equivalence between the predicted answer and the gold answer given the question, modified from QwenLong-L1~\citep{qwenlongl1}.}
\begin{tcolorbox}[
  title=\textbf{LLM Judge Prompt},
  fonttitle=\bfseries,                      
  colback=myblue!10,           
  colbacktitle=myblue!75,         
  coltitle=black,                 
  colframe=myblue!80!black,    
  coltext=black,                  
  boxrule=0.5pt,
  arc=2mm
]
You are an expert in verifying if two answers are the same.

Your input is a problem and two answers, Answer 1 and Answer 2. You need to check if they are equivalent.

Your task is to determine if two answers are equivalent, without attempting to solve the original problem.

Compare the answers to verify they represent identical values or meaning, even when written in different forms or notations.

Your output must follow the following format:

1) Provide an explanation for why the answers are equivalent or not.

2) Then provide your final answer in the form of: [[YES]] or [[NO]]

Problem: \{question\}

Answer 1: \{predicted answer\}

Answer 2: \{gold answer\}
\end{tcolorbox}
    \label{tab:llm_judge}
\end{table}


\subsection{Overall Long-context Results}

\begin{table}[!t]
\centering
    \caption{Main results across long-context reasoning benchmarks. The results for MRCR and CorpusQA\protect\footnotemark{} correspond to the 0-128K token subset. A detailed breakdown for LBV2 and LBV1-QA is provided in the Appendix~\ref{app:detailed_results_lbv1_lbv2}. We highlight the \textbf{\textcolor{red}{top-1}} and \textbf{top-3} performance. $\Delta$ indicates the performance \textcolor{blue}{gains} and \textcolor{green}{declines} compared to the base models.}\label{tab:main}
        \resizebox{0.999\textwidth}{!}{
        \begin{tabular}{lccccccc}
        \toprule
        \textbf{Models}  & \textbf{Avg.} & \textbf{DocMath} & \textbf{LBV2} & \textbf{Frames}  & \textbf{MRCR} & \textbf{CorpusQA} & \textbf{LBV1-QA} \\
        \midrule
        \multicolumn{8}{c}{\textbf{Flagship Reasoning Models}} \\ \midrule
        Gemini-2.5-Pro & \textbf{72.40}&62.38&\textbf{\textcolor{red}{65.72}}&74.51&\textbf{79.92}&\textbf{80.62}&\textbf{71.28}  \\
        GPT-5 & \textbf{\textcolor{red}{74.74}}& \textbf{\textcolor{red}{67.62}}& \textbf{62.82}& \textbf{\textcolor{red}{84.59}}& 77.29& \textbf{\textcolor{red}{81.56}}& \textbf{\textcolor{red}{73.70}} \\
        DeepSeek-R1-0528 & 68.67&63.44&\textbf{59.48}&\textbf{76.86}&64.88&77.50&69.90 \\
        Qwen3-235B-A22B-Thinking-2507 &68.45&\textbf{65.75}&57.46&75.12&66.17&75.31&\textbf{70.90} \\
        Qwen3-Max-Thinking-Preview & 69.43& 64.12&57.89&\textbf{77.93}&71.24&74.69&70.71  \\ \midrule
        \multicolumn{8}{c}{\textbf{Lightweight Reasoning Models}} \\ \midrule
        Gemini-2.5-Flash-Thinking&68.73&64.75&56.77&65.78&\textbf{78.84}&79.38&66.86    \\
        GPT-5-Nano & 57.06&63.88&43.74&73.54&43.88&50.31&67.10 \\
        GPT-OSS-120B & 58.55& 61.25&47.01&72.69&39.68&64.38&66.30 \\
        QwenLong-L1 & 56.11&64.75&40.76&72.39&47.86&42.50&68.40 \\
        Qwen3-30B-A3B-Thinking-2507 & 61.92&62.26&49.11&70.27&51.27&71.56&67.10 \\
 \midrule
        \multicolumn{8}{c}{\textbf{Ours}} \\ \midrule
        \rowcolor{Gray} \modelname-30B-A3B&\textbf{71.82}&\textbf{66.26}&55.27&74.76&\textbf{\textcolor{red}{82.99}}&\textbf{81.25}&70.40 \\
        \quad \textit{$\Delta$ to Qwen3-30B-A3B-Thinking-2507} & (\textcolor{blue}{+9.90}) & (\textcolor{blue}{+4.00}) & (\textcolor{blue}{+6.16}) & (\textcolor{blue}{+4.49}) & (\textcolor{blue}{+31.72}) & (\textcolor{blue}{+9.69}) & (\textcolor{blue}{+3.30}) \\
        \bottomrule
    \end{tabular}
    }
\end{table}

\footnotetext{Unless specified otherwise, "MRCR" and "CorpusQA" refers to the 0-128K token subsets. Subsets with longer contexts are explicitly labeled in subsequent analyses (e.g., Table~\ref{tab:memory}).}

To evaluate the long-context reasoning capabilities of \modelname, we conducted a comprehensive  analysis as detailed in Table~\ref{tab:main}. We benchmark our model, \modelname-30B-A3B, against the leading flagship reasoning models such as GPT-5 and Gemini-2.5-Pro, and various lightweight reasoning models, including our direct baseline, Qwen3-30B-A3B-Thinking-2507. There are several observations:

\paragraph{Overall Performance Evaluation} \modelname-30B-A3B achieves an average score of 71.82 across the evaluated benchmarks. This performance demonstrates a clear advantage over other prominent models. Specifically, it surpasses leading open-source reasoning models such as DeepSeek-R1-0528 (68.67), strong lightweight models like Gemini-2.5-Flash-Thinking (68.73), and shows substantial improvement over our baseline, Qwen3-30B-A3B-Thinking-2507 (61.92). Furthermore, \modelname-30B-A3B's performance is competitive with top-tier flagship models, approaching the score of Gemini-2.5-Pro (72.40) and it achieves the state-of-the-art score of 82.99 on the MRCR benchmark. These results collectively indicate that our proposed methodology enables a 30B-A3B model to reach a level of performance previously associated with much larger-scale systems.

\paragraph{Analysis of Performance on Specific Task Categories} An analysis of the results indicates that the \modelname-30B-A3B's performance gains are most pronounced in tasks requiring complex reasoning and information integration. This observation is consistent with the objectives of our synthetic data generation strategy. We identify two task categories where our model shows notable performance:
\begin{itemize}
    \item \textbf{Multi-hop Reasoning:} On benchmarks such as Longbench-V2, Frames, and LongBench-V1-QA, which require connecting discontinuous information to form logical chains, our model achieves performance comparable to that of the flagship models and exceeds that of other lightweight models.
    \item \textbf{Information Aggregation and Intensive Calculation:} In tasks like CorpusQA, which require aggregating scattered information from the context for subsequent calculation or synthesis, \modelname-30B-A3B scores 81.25. This is competitive with the score of GPT-5 (81.56), suggesting a high proficiency in processing distributed information.
\end{itemize}
This performance pattern, with heightened scores on complex reasoning tasks, suggests that our data synthesis pipeline is effective in enhancing the model's ability to perform multi-hop reasoning among the long-context, as opposed to simple information retrieval.

\paragraph{Quantifying the Impact on Long-Context Performance} The contribution of our methodology is further quantified by a direct comparison with the baseline model, Qwen3-30B-A3B-Thinking-2507. Our \modelname-30B-A3B obtains a +9.90 point improvement in the average score. An important observation is that the largest performance gains are concentrated on benchmarks characterized by longer average context lengths. Specifically, the most substantial gains are on MRCR (+31.72, avg. 36.5K tokens\footnote{The number of tokens was computed using the Qwen3-30B-A3B-Thinking-2507 tokenizer.}), CorpusQA (+9.69, avg. 92.8K tokens), and LongBench-V2 (+6.16, avg. 85.5K tokens). This correlation between performance improvements and longer contexts suggests that our proposed methodology is particularly effective at addressing the challenges inherent to reasoning over extended context lengths.

\subsection{Generalization Benefits from Long-Context Enhancement}

\begin{figure}[!t]
    \centering
    \begin{minipage}[t]{0.5\textwidth}
        \centering
        \captionof{table}{Comparison of Qwen3-30B-A3B-Thinking-2507 and \modelname on general, agentic memory, and dialogue memory benchmarks.}
        \label{tab:res_general_vertical}
        \setlength{\tabcolsep}{4pt}
        \small 
        \begin{tabular}{lcc}
            \toprule
            \textbf{Benchmark} & \textbf{\shortstack{Qwen3-30B-A3B-\\Thinking-2507}} & \textbf{\shortstack{QwenLong-L1.5\\-30B-A3B}} \\
            \midrule
            \multicolumn{3}{c}{\textit{general}} \\
            \midrule
            \quad MMLU-PRO & 81.03 & 81.33 (\textcolor{blue}{+0.30}) \\
            \quad AIME24 & 90.31 & 90.0 (\textcolor{green}{-0.31}) \\
            \quad AIME25 & 82.81 & 86.46 (\textcolor{blue}{+3.65}) \\
            \quad GPQA-Diamond & 75.88 & 76.78 (\textcolor{blue}{+0.90}) \\
            \midrule
            \multicolumn{3}{c}{\textit{agentic memory}\ (BFCL-V4)} \\
            \midrule
            \quad Memory-Sum & 23.01 & 24.52 (\textcolor{blue}{+1.51}) \\
            \quad Memory-KV & 10.97 & 16.77 (\textcolor{blue}{+5.80}) \\
            \quad Memory-Vec & 16.13 & 16.77 (\textcolor{blue}{+0.64}) \\
            \quad Memory-Rec\_Sum & 41.94 & 40.00 (\textcolor{green}{-1.94}) \\
            \midrule
            \multicolumn{3}{c}{\textit{dialogue memory}} \\
            \midrule
            \quad LongMemEval & 60.80 & 76.40 (\textcolor{blue}{+15.60}) \\
            \bottomrule
        \end{tabular}
    \end{minipage}%
    \hfill
    \begin{minipage}[t]{0.45\textwidth}
        \centering
        \vspace*{14pt} 
        
        \includegraphics[width=\linewidth]{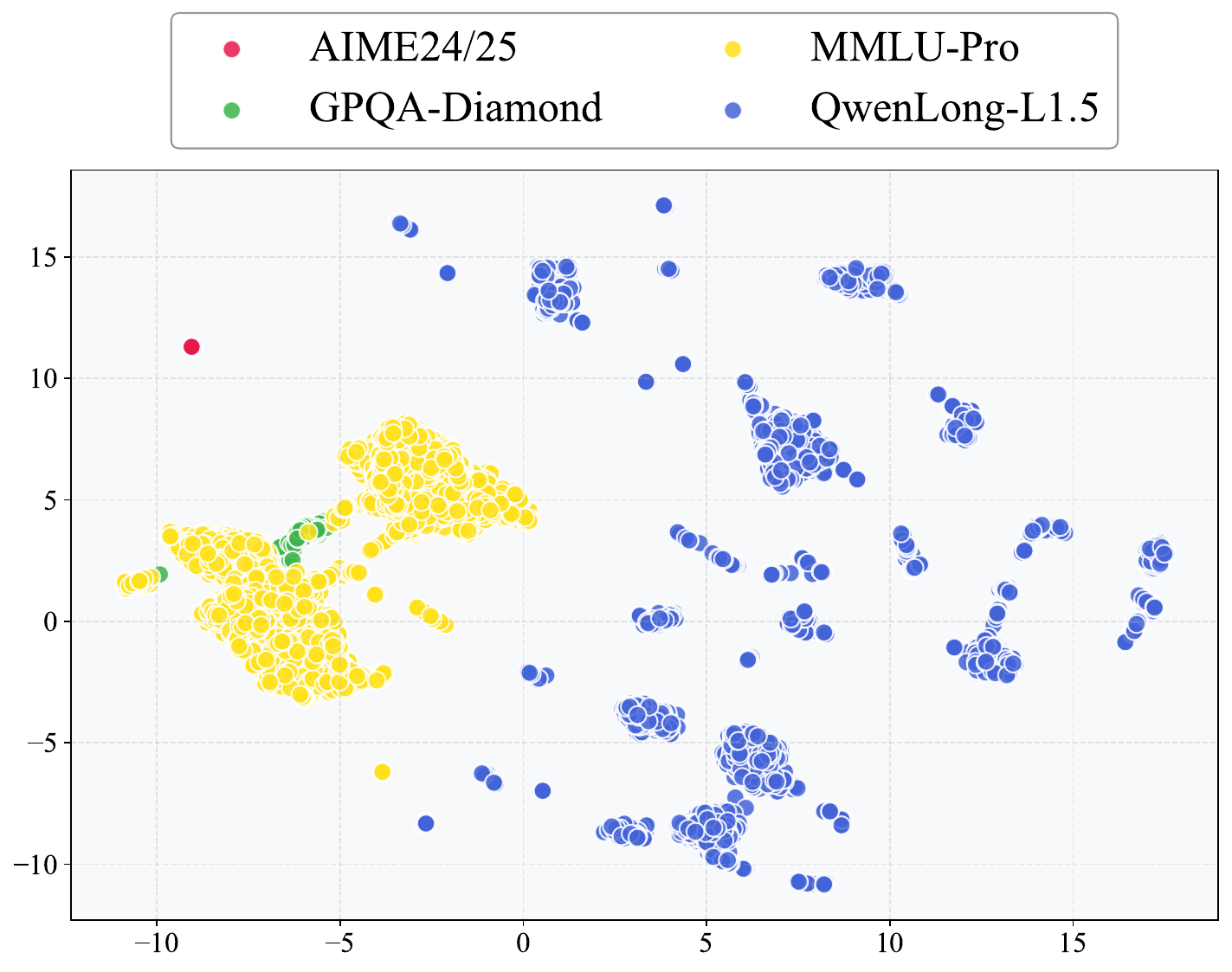}
        
        \captionof{figure}{Two-dimensional UMAP projection of AIME24/25, GPQA-Diamond, MMLU-Pro, and QwenLong-L1.5 training set.}
        \label{fig:data_vis_testset}
    \end{minipage}
\end{figure}


To assess whether the skills acquired during long-context post-training generalize to other domains, we compare our model \modelname-30B-A3B against its baseline, Qwen3-30B-A3B-Thinking-2507. As shown in Table~\ref{tab:res_general_vertical}, our evaluation spans three areas beyond our primary training focus. General capabilities are measured using a suite of standard benchmarks including MMLU-PRO~\citep{wang2024mmlupro}, AIME24/25~\citep{AIME} and GPQA-Diamond~\citep{rein2023gpqagraduatelevelgoogleproofqa}. Agentic memory is evaluated with the BFCL-V4 memory subset~\citep{patil2025bfcl} and dialogue memory is assessed using LongMemEval~\citep{wu2024longmemeval}. The results indicate that our methodology induces positive generalization rather than catastrophic forgetting.

The evaluation first reveals that our model avoids performance degradation on general knowledge and reasoning tasks. This stability is particularly noteworthy given that these benchmarks are evaluated out-of-distribution. As visually confirmed in Figure~\ref{fig:data_vis_testset}, our training data occupie a distinct semantic space from these evaluation sets. We attribute this robust generalization to our RL strategy, which effectively regularizes the training process and prevents the catastrophic forgetting of foundational abilities. Beyond merely retaining performance, \modelname-30B-A3B shows gains on reasoning-intensive tasks such as AIME25 (+3.65) and GPQA-Diamond (+0.90). This suggests that the information integration skills from long-context training are transferable, enhancing the model's ability to maintain focus and integrate key information within its own long-form reasoning outputs required by these tasks. 
We provide a case study in Appendix~\ref{appendix:aime_case} to illustrate this improved reasoning capabilities. 

This principle of generalizing core information-integration skills is further demonstrated in the domain of agentic memory. \modelname-30B-A3B shows improved performance across most sub-tasks of the BFCL-V4 benchmark, with a particularly notable gain on Memory-KV (+5.80). This indicates that the ability to identify and analyze key information within long documents is transferable to managing the structured, sequential history of an agent's operations, thereby enhancing its overall effectiveness.

Finally, the effectiveness of our training is most directly validated in the domain of dialogue memory. The model achieves a substantial +15.60 point gain on the LongMemEval benchmark. As maintaining state and recalling information over extended conversations is a primary application for long-context models, this result directly confirms the success of our methodology in its intended domain. Collectively, these findings suggest that enhancing long-context processing through our proposed method leads to broad and fundamental improvements in a model's cognitive abilities.

\subsection{Length Extension with Memory Management}

\begin{table}[!t]
\centering
    \caption{Results on the subsets MRCR and CorpusQA with over 128K tokens lengths.}\label{tab:memory}
        \resizebox{0.7\textwidth}{!}{
        \begin{tabular}{lcccc}
        \toprule
        \multirow{2}{*}{\textbf{Models}}&\multicolumn{2}{c}{\textbf{MRCR}}&\multicolumn{2}{c}{\textbf{CorpusQA}}\\
          & \textbf{128K$\sim$512K} & \textbf{512K$\sim$1M} & \textbf{1M}  & \textbf{4M}   \\
        \midrule
        \multicolumn{5}{c}{\textbf{Full-context Inference}} \\ \midrule
        Gemini-2.5-Pro & 53.83&39.51&53.11&- \\
        Gemini-2.5-Flash-Thinking & 53.98& 46.88& 36.91& - \\
        Qwen-Flash-Thinking-1M &22.10 & 9.97& 2.75&- \\
        \midrule
        \multicolumn{5}{c}{\textbf{Memory Agent Framework}} \\ \midrule
        MemAgent-14B~\citep{yu2025memagent}& 6.78&3.11 & 9.70 & 9.09 \\
        Qwen3-30B-A3B-Thinking-2507 & 16.55 &4.24 & 15.32 & 9.52 \\
        \rowcolor{Gray}QwenLong-L1.5-30B-A3B &  34.87& 22.53 & 20.72 & 14.29 \\
        \bottomrule
    \end{tabular}
    }
\end{table}

We evaluate ultra-long context performance on challenging benchmarks. Specifically, we use subsets of MRCR with contexts over 128K tokens to test complex retrieval via its rich content and positional sorting, alongside CorpusQA to measure multi-hop grounding over scattered information, with some instances reaching up to 4 million tokens. We compare \modelname-30B-A3B operating within our memory agent framework against other agent-based methods and leading full-context models.

The results in Table~\ref{tab:memory} show that within the memory agent framework,  \modelname-30B-A3B demonstrates a performance advantage over its peers. On the MRCR (128K~512K) subset, it achieves a score of 34.87, which is 18.32 points higher than  Qwen3-30B-A3B-Thinking-2507. This performance margin is maintained as the context length increases; in the 512K$\sim$1M range, \modelname-30B-A3B  scores 22.53, again more than 18 points higher than Qwen3-30B-A3B-Thinking-2507. This consistent performance gap highlights the effectiveness of learned memory compression and planning ability of \modelname-30B-A3B.

The scalability of the memory agent framework becomes particularly evident at the 4M token scale, a context length that is intractable for current full-context methods. On the CorpusQA 4M token subset, \modelname-30B-A3B achieves a score of 14.29, demonstrating its ability to perform reasoning at extreme scales. This result underscores the framework's advantage in handling tasks that lie beyond the operational limits of LLMs, showcasing its robustness and extensibility.

The results confirm that a leading proprietary model like Gemini-2.5-Pro currently exhibits the strongest performance on these ultra-long context tasks. Within this competitive landscape, our work validates the effectiveness of memory agent framework. \modelname-30B-A3B outperforms other agent-based methods and surpasses select full-context models such as Qwen-Flash-Thinking-1M on these challenging benchmarks. Furthermore, the results indicate that our approach has favorable context scaling properties. We consider this a robust foundation and will focus on further optimization in our future work.

\subsection{Performance Dynamics of Progressive Long-context Post-training}\label{sec:exp_multi_stage}

\begin{table}[!t]
\centering
    \caption{Performances of \modelname-30B-A3B after different post-training stages.}\label{tab:res_stages_long}
    \resizebox{0.999\textwidth}{!}{
    \begin{tabular}{l|ccccccc|c}
    \toprule
    \multirow{2}{*}{\textbf{Models}}&\multicolumn{7}{c|}{\textbf{Full-context Evaluation}} & \textbf{Memory Agent} \\
    & \textbf{Avg.} & \textbf{DocMath} & \textbf{LBV2} & \textbf{Frames}  & \textbf{MRCR} & \textbf{CorpusQA} & \textbf{LBV1-QA} & \textbf{MRCR (512K$\sim$1M)} \\
    \midrule
    Qwen3-30B-A3B-Thinking-2507 & 61.92 & 62.26 & 49.11 & 70.27 & 51.27 & 71.56 & 67.10& 4.24 \\
    \quad\quad+Naive GRPO  & 67.24 & 65.12 & 55.27 & 71.36 & 66.92 & 76.87 & 67.90& - \\
    \midrule
    \modelname-30B-A3B& &  &  &  &  &  & &\\
    \quad\quad $\vdash$ Full-context RL Stage-1 & 69.59 & 65.88 & 55.67 & 73.79 & 76.35  & 75.62 & 70.20 & 17.14 \\
    \quad\quad $\vdash$ Full-context RL Stage-2 & 70.46 & 64.38 & 53.68 & 74.76 & 81.53 &  77.50 & 70.90& 17.05 \\
    \quad\quad $\vdash$ Full-context RL Stage-3 & 71.59 & 66.25 & 57.65 & 74.27 & 82.69  & 79.38 & 69.30 &12.66 \\
    \quad\quad $\vdash$ Memory-RL & 68.53 & 64.75&47.91&74.15&80.29&75.00&69.10&20.34\\
    \quad\quad $\vdash$ Model Merging w/ SCE & 71.18 & 66.75 & 52.88 & 73.79 & 82.69  & 81.56 & 69.40 &21.68\\
    \quad\quad $\vdash$ Full-context RL Stage-4 & 71.82 & 66.26 & 55.27 & 74.76 & 82.99 & 81.25 & 70.40 &22.53 \\
    \bottomrule
    \end{tabular}
    }
\end{table}

To analyze the contributions of our multi-stage post-training strategy, we present an ablation study in Table~\ref{tab:res_stages_long}. The results show a consistent improvement in the average score as training progresses through the stages, and the final version of \modelname-30B-A3B (71.82) outperforms the baseline trained with a naive GRPO approach (67.24), validating the effectiveness of our proposed RL strategy in Section~\ref{sec:rl_method}. Notably, a substantial portion of the performance gain is achieved after just the first stage (Full-context RL Stage-1), where the average score increases from 61.92 to 69.59. This demonstrates that the synthetic data used in Stage-1 activates the model's foundational ability to ground and reason over long documents.

A closer analysis of the full-context evaluation benchmarks reveals how different stages impact specific capabilities. On benchmarks with shorter average input lengths and lower information density, such as DocMath, Frames, and LBV1-QA, performance remains stable after the initial boost from Stage-1. In contrast, for tasks characterized by longer contexts and the need for global information aggregation, such as MRCR and CorpusQA, performance consistently improves as training progresses from Stage 1 to 3 (e.g., MRCR score rises from 76.35 to 82.69). This distinction highlights the necessity of our progressive length extension strategy for developing the advanced, full-context reasoning skills required by the most information-intensive tasks.

The evolution of the model's memory agent capability on MRCR-1M provides further insights into our methodology. While the first RL stage provides a significant boost over the baseline, the results show that a specialized memory-RL training stage is necessary to achieve further gains in this area. However, this specialization leads to a trade-off, as the model's full-context inference performance drops to 68.53 after this stage. 
The model merging stage effectively resolves this conflict by restoring the full-context capability to 71.18 while further improving the memory agent score to 21.68. Interestingly, the acquired memory skill demonstrates strong robustness; subsequent full-context-RL training in fourth stage does not degrade this specialized ability but instead further enhances it to 22.53. Ultimately, this multi-stage process yields a single, unified model proficient in both direct long-context reasoning via full-context inference and in functioning as a memory-augmented agent.





\section{Conclusion}
In this work, we introduced \modelname, a comprehensive post-training recipe that  elevates long-context reasoning to the level of top-tier proprietary models.  Our core contribution is a full post-training system that unifies a scalable data synthesis pipeline, tailored RL methodologies, and a memory-augmented agent architecture. Our comprehensive evaluation reveals that \modelname achieves performance comparable to flagship models like GPT-5 and Gemini-2.5-Pro, with significant gains on tasks requiring multi-hop reasoning and information aggregation. We observed that these performance improvements progressively scale with context length and complexity, validating the effectiveness of our data synthesis and RL strategies. Crucially, our findings indicate that the sophisticated reasoning skills developed during long-context training are not isolated, they generalize effectively, enhancing performance on out-of-domain benchmarks in mathematics, tool-use agents, and long-form dialogue. Furthermore, the integration of our memory management framework, refined through a multi-stage fusion paradigm, extends the model's capabilities to the 1M$\sim$4M token regime, successfully unifying its single-pass reasoning prowess and its iterative memory-agent abilities into a single, cohesive model. Collectively, these results demonstrate that our approach, unifying data, training, and agent framework, provides a robust and scalable pathway for advancing the long-context reasoning capabilities of open LLMs.

\section{Limitations and Future Works}
While \modelname demonstrates significant progress in long-context reasoning, we identify several avenues for future research. These opportunities are primarily centered on expanding our data synthesis system and refining the reinforcement learning framework.

\subsection{Data Coverage and Synthesis}

\paragraph{Scaling Data Diversity and Task Complexity:} Real-world applications often involve not only long inputs but also long-form outputs, such as chapter-level document revision, report generation from source materials, and complex tasks orchestrated by autonomous agents. Our current data synthesis pipeline is not yet optimized for these scenarios. A key future direction is to expand our task taxonomy to cover these long-input, long-output problems. Furthermore, our current data is exclusively text-based. We plan to extend our framework to incorporate multi-modal data, which presents a significant and exciting challenge for sequential reasoning.

\paragraph{Developing a Closed-Loop Data Flywheel:} Although our data synthesis pipeline is automated, its scalability is constrained by practical bottlenecks, namely the API quotas of proprietary models and the computational cost of serving large open-source models for generating long-context data. To mitigate these resource dependencies, we propose developing a closed-loop data flywheel. The core idea is to leverage a model, once it has been enhanced by our training method, to become a data generator itself. This powerful long-context model could then be used to create vast quantities of new QA pairs and, critically, their corresponding thinking trajectories. Such a self-improving loop would substantially reduce the reliance on external resources and accelerate data scaling in a cost-effective manner.

\subsection{Reinforcement Learning Algorithm}

\paragraph{Granular Credit Assignment for Thinking Trajectories:} As discussed in Section~\ref{sec:token-level-ngc}, we identified that GRPO's reward assignment can lead to training instability. While our proposed AEPO method effectively mitigates this issue through techniques like gradient clipping, it serves as a robust stabilization mechanism rather than a fundamental solution to the credit assignment problem. The core challenge remains: our current approach assigns a uniform advantage signal to an entire reasoning step. A primary focus of our future research will be to develop a token-level credit assignment mechanism within the thinking trajectory. This would allow the model to differentiate the contribution of each token within a generated thought or plan, enabling more precise and efficient learning.

\paragraph{Sophistication of the Reward Model:} Our current implementation relies on a reward function combining simple rule-based checks and an LLM-as-a-judge paradigm. This approach is effective for tasks with clear-cut correctness criteria, such as question answering. However, its efficacy diminishes for more open-ended and subjective tasks where "correctness" is multifaceted. To address this, a significant future direction is the research and development of more sophisticated reward systems. Specifically, we aim to explore LLM-based rubric reward models, where a model is trained to score responses against a detailed, multi-faceted rubric. Such models could provide richer, more nuanced reward signals, better aligning the agent's behavior with complex human preferences in real-world scenarios.

\clearpage
\bibliography{iclr2024_conference}
\bibliographystyle{iclr2024_conference}

\clearpage
\appendix

\section{Detailed Results on LongBench-V2 and LongBench-V1 QA Subsets}\label{app:detailed_results_lbv1_lbv2}
\begin{table}[!h]
\centering
    \caption{Detailed results on LongBench-V2 subsets. $\Delta$ indicates the performance \textcolor{blue}{gains} compared to the base model.}
    \label{tab:appendix_lbv2}
    \resizebox{0.9\textwidth}{!}{
    \begin{tabular}{lcccccc}
    \toprule
    \multirow{2}{*}{\textbf{Models}} & \multicolumn{6}{c}{\textbf{LongBench-V2}} \\
    \cmidrule(lr){2-7}
    & \textbf{Overall} & \textbf{Easy} & \textbf{Hard} & \textbf{Short} & \textbf{Medium} & \textbf{Long} \\
    \midrule
    \multicolumn{7}{c}{\textbf{Flagship Reasoning Models}} \\
    \midrule
    Gemini-2.5-Pro & 65.72 & 72.34 & 61.64 & 69.66 & 62.91 & 64.71 \\
    DeepSeek-R1-0528 & 59.48 & 65.61 & 55.70 & 66.11 & 52.86 & 61.32 \\
    \midrule
    \multicolumn{7}{c}{\textbf{Lightweight Reasoning Models}} \\
    \midrule
    Gemini-2.5-Flash-Thinking & 56.77 & 67.19 & 50.32 & 61.67 & 52.80 & 56.48 \\
    Qwen3-30B-A3B-Thinking-2507 & 49.11 & 52.60 & 46.95 & 59.44 & 43.72 & 42.59 \\
    \midrule
    \multicolumn{7}{c}{\textbf{Ours}} \\
    \midrule
    \rowcolor{Gray} \modelname-30B-A3B & 55.27 & 59.90 & 52.41 & 62.78 & 53.95 & 45.37 \\
    \quad \textit{$\Delta$ to Qwen3-30B-A3B-Thinking-2507} & (\textcolor{blue}{+6.16}) & (\textcolor{blue}{+7.30}) & (\textcolor{blue}{+5.49}) & (\textcolor{blue}{+3.34}) & (\textcolor{blue}{+10.23}) & (\textcolor{blue}{+2.87}) \\
    \bottomrule
    \end{tabular}
    }
\end{table}

\begin{table}[!h]
\centering
    \caption{Detailed results on LongBench-V1 QA subsets. $\Delta$ indicates the performance \textcolor{blue}{gains} and \textcolor{green}{declines} compared to the base model.}
    \label{tab:appendix_lbv1_qa}
    \resizebox{0.9\textwidth}{!}{
    \begin{tabular}{lcccccc}
    \toprule
    \multirow{2}{*}{\textbf{Models}} & \multicolumn{6}{c}{\textbf{LongBench-V1 QA}} \\
    \cmidrule(lr){2-7}
    & \textbf{Avg.} & \textbf{2Wiki.} & \textbf{HotpotQA} & \textbf{Musique} & \textbf{NarrativeQA} & \textbf{Qasper} \\
    \midrule
    \multicolumn{7}{c}{\textbf{Flagship Reasoning Models}} \\
    \midrule
    Gemini-2.5-Pro & 71.28 & 91.00 & 82.41 & 71.50 & 61.00 & 50.50 \\
    DeepSeek-R1-0528 & 69.90 & 90.00 & 82.00 & 68.50 & 61.00 & 48.00 \\
    \midrule
    \multicolumn{7}{c}{\textbf{Lightweight Reasoning Models}} \\
    \midrule
    Gemini-2.5-Flash-Thinking & 66.86 & 89.34 & 81.00 & 60.00 & 56.50 & 47.50 \\
    Qwen3-30B-A3B-Thinking-2507 & 67.10 & 88.00 & 77.50 & 64.50 & 52.00 & 53.50 \\
    \midrule
    \multicolumn{7}{c}{\textbf{Ours}} \\
    \midrule
    \rowcolor{Gray} \modelname-30B-A3B & 70.40 & 91.00 & 82.41 & 71.50 & 61.00 & 50.50 \\
    \quad \textit{$\Delta$ to Qwen3-30B-A3B-Thinking-2507} & (\textcolor{blue}{+3.30}) & (\textcolor{blue}{+3.00}) & (\textcolor{blue}{+4.91}) & (\textcolor{blue}{+7.00}) & (\textcolor{blue}{+9.00}) & (\textcolor{green}{-3.00}) \\
    \bottomrule
    \end{tabular}
    }
\end{table}

To further analyze the performance of \modelname, we provide a detailed breakdown of its results on the LongBench-V2 and LongBench-V1 QA subsets in Table~\ref{tab:appendix_lbv2} and Table~\ref{tab:appendix_lbv1_qa}. This fine-grained analysis indicates that the performance gains are concentrated in specific areas that align with our post-training strategy.

On LongBench-V2, \modelname shows performance increases over its baseline across all sub-dimensions, including difficulty and context length. The largest gain is on the Medium length subset (+10.23). This subset corresponds to the 32K$\sim$128K token range, a focus area for our synthetic data generation. This correlation suggests that our targeted data synthesis pipeline is effective at improving performance within this context windows.

The results on LongBench-V1 QA show that \modelname achieves performance comparable to Gemini-2.5-Pro on several key reasoning tasks. The +7.00 point improvement on Musique, a benchmark designed for multi-hop reasoning, points to the effectiveness of our data synthesis pipeline. Similarly, the +9.00 point gain on NarrativeQA, which has one of the longest contexts in the LongBench-V1 suite, indicates an improved capacity for information integration over extended documents. While a small performance decrease is observed on Qasper (-3.00), the overall trend supports the conclusion that our methodology enhances both specific reasoning patterns and general long-context processing.

\section{QwenLong-L1.5 Synthetic Data Cases}

In this section, we list the detailed cases of RL training data we constructed during long-context data synthetic pipeline, including \textbf{Numerical Calculation}, \textbf{Long In-context Learning}, \textbf{Viewpoint Analysis}, \textbf{Multi-fact Reasoning}, \textbf{Hypothetical Scenarios}, \textbf{Temporal Reasoning}, \textbf{Causal Analysis} and e.t.c.

\begin{tcolorbox}[colback=white, colframe=gray!50!black, title=Question Type: \textbf{Numerical Calculation}]
    \textbf{Question:}\\
    Based on the CECONY financial data for the twelve months ended December 31, what is the difference, in millions of dollars, between the total operating expenses for the Electric segment in 2013 and the total operating expenses for the Gas segment in 2014? Define total operating expenses as the sum of all itemized costs listed between 'Operating revenues' and 'Operating income'.
    
    \vspace{2mm}
    \textbf{Answer:}\\
    5129

    \vspace{2mm}
    \tcblower
    \textbf{Document:}\\
    ABIOMED, INC.  AND SUBSIDIARIES Notes to Consolidated Financial Statements—(Continued) Note 12.\\
Stock Award Plans and Stock Based Compensation (Continued) Restricted Stock The following table summarizes restricted stock activity for the fiscal year ended March 31, 2009:\\
|| March 31, 2009|\\
||Number of Shares (in thousands)| Grant Date Fair Value|\\
|Restricted stock awards at March 31, 2008|54|\$11.52|\\
|Granted|666|16.75|\\
|Vested|-167|14.65|\\
|Forfeited|-73|17.53|\\
|Restricted stock awards at March 31, 2009|480|\$16.77| \\
The remaining unrecognized compensation expense for restricted stock awards at March 31, 2009 was \$4.6 million.\\
The weighted average remaining contractual life for restricted stock awards at March 31, 2009 and 2008 was 1.8 and 2.4 years, respectively.\\
In May 2008, 260,001 shares of restricted stock were issued to certain executive officers and certain members of senior management of the Company, of which 130,002 of these shares vest upon achievement of a prescribed performance milestone.\\
In September 2008, the Company met the prescribed performance milestone, and all of these performance-based shares vested.\\
In connection with the vesting of these shares, these employees paid withholding taxes due by returning 39,935 shares valued at \$0.7 million.\\
These shares have been recorded as treasury stock as of March 31, 2009.\\
The remaining 129,999 of the restricted shares award vest ratably over four years from the grant date.\\

    \textcolor[RGB]{158,160,161}{[Intermeidate document text abbreviated]}\\

    This structure is the basis for our reportable segment information discussed below.\\
Management evaluates operating segment performance based upon segment operating profit exclusive of operating expenses pertaining to global operations and corporate expenses, share-based compensation expense, settlement, certain claims, acquisition, integration and other expenses, inventory step-up, in-process research and development write-offs and intangible asset amortization expense.\\
Global operations include research, development engineering, medical education, brand management, corporate legal, finance, and human resource functions, and U. S.  and Puerto Rico-based manufacturing operations and logistics.\\
Intercompany transactions have been eliminated from segment operating profit.\\
Management reviews accounts receivable, inventory, property, plant and equipment, goodwill and intangible assets by reportable segment exclusive of U.\\
S and Puerto Rico-based manufacturing operations and logistics and corporate assets.
\end{tcolorbox}

\begin{tcolorbox}[colback=white, colframe=gray!50!black, title=Question Type: \textbf{Long In-context Learning}]
    \textbf{Question:}\\
A developer is quantizing a custom Llama-like model with a dimension (`dim') of 3584 and a feed-forward intermediate size (`hidden\_size') of 9984. They first convert the Hugging Face checkpoint using `scripts/convert\_hf\_checkpoint.py', then quantize the resulting `model.pth` by running `python quantize.py --mode int4-gptq --groupsize 256'. Considering the logic in `convert\_hf\_checkpoint\_llama', the `WeightOnlyInt4GPTQQuantHandler' implementation, and the `WeightOnlyInt4Linear' module definition (assuming default `inner\_k\_tiles=8' and `padding=True'), what are the exact final shapes of the `weight' and `scales\_and\_zeros' tensors for the following two layers as they are saved in the final quantized checkpoint?\\
1. The combined attention projection layer (`wqkv').\\
2. The feed-forward down-projection layer (`w2').\\
Assume `find\_multiple(k, n)' is a function that returns the smallest integer $\geq$ k that is a multiple of n.

    \begin{verbatim}
(A) 1. For the `wqkv` layer:
    - `weight` shape: `(1344, 28, 32, 4)`
    - `scales_and_zeros` shape: `(14, 10752, 2)`

    2. For the `w2` layer:
    - `weight` shape: `(448, 78, 32, 4)`
    - `scales_and_zeros` shape: `(39, 3584, 2)`
(B) 1. For the `wqkv` layer:
    - `weight` shape: `(448, 32, 32, 4)`
    - `scales_and_zeros` shape: `(16, 3584, 2)`

    2. For the `w2` layer:
    - `weight` shape: `(448, 80, 32, 4)`
    - `scales_and_zeros` shape: `(40, 3584, 2)`
(C) 1. For the `wqkv` layer:
    - `weight` shape: `(1344, 32, 32, 4)`
    - `scales_and_zeros` shape: `(16, 10752, 2)`

    2. For the `w2` layer:
    - `weight` shape: `(448, 80, 32, 4)`
    - `scales_and_zeros` shape: `(40, 3584, 2)`
(D) 1. For the `wqkv` layer:
    - `weight` shape: `(672, 32, 32, 4)`
    - `scales_and_zeros` shape: `(10752, 16, 2)`

    2. For the `w2` layer:
    - `weight` shape: `(224, 80, 32, 4)`
    - `scales_and_zeros` shape: `(3584, 40, 2)`
\end{verbatim}
    
    \vspace{2mm}
    \textbf{Answer:}\\
    (C)

    \vspace{2mm}
    \tcblower
    \textbf{Documents:}
    
    \begin{verbatim}
### FILE: scripts/prepare.sh
python scripts/download.py --repo_id $1

### FILE: mixtral-moe/scripts/convert_hf_checkpoint.py

import glob
import json
import re
import sys
from pathlib import Path
from typing import Optional
import torch

wd = Path(__file__).parent.parent.resolve()
sys.path.append(str(wd))
\end{verbatim}

    \textcolor[RGB]{158,160,161}{[Following document text abbreviated]}
\end{tcolorbox}

\begin{tcolorbox}[colback=white, colframe=gray!50!black, title=Question Type: \textbf{Viewpoint Analysis}]
    \textbf{Question:}\\
    What is the correct answer to this question:  When Sophia suggests that Brenda and Lawrence might have conspired to kill Aristide, Charles finds the inference 'a bit far-fetched' because of the following reasons:\\
Choices:\\
(A)  Brenda sympathizes with Lawrence.\\
(B)  Brenda has a psychological desire for wealth and a comfortable life.\\
(C)  Sophia's family treated Brenda unfairly.\\
(D)  Brenda is very afraid.\\
    
    \vspace{2mm}
    \textbf{Answer:}\\
    (C)

    \vspace{2mm}
    \tcblower
    \textbf{Document:}\\
    "Who had the motive?"
"No one - no one would want to kill him!"\\
"Would you like a lawyer present?" Taverna asked.\\
"I don't have one. I don't need one. I have nothing to hide - nothing..."\\
"You should know that what you say can be used as evidence?"\\
"I'm innocent. I swear - I'm innocent."\\
"I didn't imply anything."\\
At this point, Taverna paused and changed the topic. "Leonidis' wife is much younger than her husband, isn't she?"\\
"I - I think so - I mean, they are quite a few years apart."\\
"She must feel terribly lonely sometimes."\\
Lawrence Brown did not respond, merely licking his dry lips with his tongue.\\
"Having a companion her own age would surely make her very happy?"\\
"I—I'm not—I mean—I don't know."\\
"In my opinion, it's only natural for the two of you to develop a reliance on each other."\\
The young man protested vehemently.\\
"No, not at all! There's nothing like that! I know exactly what you're thinking, but there's nothing like that! Mrs. Lionides has always been very kind to me – I greatly respect her – but that's all – nothing more – too absurd, too absurd! I wouldn't kill anyone – much less do something like substitution. I'm very sensitive and easily excitable. I – I can't even conceive of having murderous thoughts – the assigner understands that well – my religion forbids killing. They let me work in the hospital boiler room – the work was too tiring – I couldn't keep up – so they let me be a tutor. I did my best to teach Eustace and Josephine – Josephine was very clever, but a bit difficult to teach. Everyone here has been very kind to me – Mr. Lionides, Mrs. Lionides, and Miss Edith de Havran are all good people. Now this terrible thing has happened … and you suspect me!”\\
The Taverner inspector looked at him with a detached expression.\\
"I didn't say that," he told Brown.\\
"But you thought it. I know you thought it! They all think it! I can see it in their eyes. I – I can't continue talking to you. I feel uncomfortable."\\
He hurried out of the reading room. Taverna slowly turned his head and looked at me.\\
"What do you think of him?"\\
"He's scared," I replied.\\
"I mean, do you think he's the killer?"\\
"If you ask me," Detective Lam interrupted, "I'd say he doesn't have the guts."\\
"He wouldn't hit someone on the head or shoot them," the Chief Inspector chimed in, "but this kind of crime should be doable, right? Just mess around with a few pill bottles... it's just helping an old man leave this world painlessly."\\
"A simple, practical method of euthanasia!" Detective Inspector Lamb commented, "After the dust settles, he might even be able to marry a woman who inherits a hundred thousand pounds tax-free. The woman already has about that amount in assets, along with some large red and blue sapphires. It's definitely worth a try."
"But this is just speculation and conjecture!" Taverner sighed, "I did try to scare him, but that proves nothing. Even if he's innocent, he would still be scared like this. In fact, I don't think he did it. I'm more suspicious of the woman—but I don't know why she didn't throw away or wash the insulin bottle."
"The housekeeper said they were very close."\\
"Any evidence?"\\
"She judged it from the way Leonidis looked at Mrs. Leonidis when she served him coffee."\\
"This can't be taken to court! Is there anything else?"\\
    \textcolor[RGB]{158,160,161}{[Following document text abbreviated]}
\end{tcolorbox}

\begin{tcolorbox}[colback=white, colframe=gray!50!black, title=Question Type: \textbf{Multi-fact Reasoning}]
    \textbf{Question:}
    
    Identify the character from a separate gaming franchise whose outfit is available as an unlockable cosmetic within the final *Assassin's Creed* installment for which one of the original game's creators served as creative director. The direct predecessor to this installment is a game that can be seen being played by characters within a 2014 Ubisoft title set in Chicago. This Chicago-based game, in turn, features a side mission involving the fictional CEO of Abstergo Entertainment, a character who was first introduced in a title centered on the Golden Age of Piracy, the narrative of which precedes the events of the game that was released concurrently with the adventure of Arno Dorian.
    
    \vspace{2mm}
    \textbf{Answer:}
    Raiden.

    \vspace{2mm}
    \textbf{Reasoning Chain:} (Arno Dorian)-[is protagonist of]-(Assassin's Creed Unity)-[released concurrently with]-(Assassin's Creed Rogue)-[is set between events of]-(Assassin's Creed IV: Black Flag)-[is introduced in]-(Olivier Garneau)-[features side mission about]-(Watch Dogs)-[is played by characters in]-(Assassin's Creed II)-[is sequel to]-(Assassin's Creed)-[created by]-(Patrice Désilets)-[was creative director of]-(Assassin's Creed: Brotherhood)-[includes outfit of]-(Raiden)

    \tcblower
    \textbf{Documents:}
    
    \textit{Doc 1:}\\
    TITLE: Assassin's Creed\\
CONTENT: Assassin's Creed is a historical action-adventure video game series and media franchise published by Ubisoft and developed mainly by its studio Ubisoft Montreal using the game engine Anvil and its more advanced derivatives. Created by Patrice Désilets, Jade Raymond, and Corey May, the Assassin's Creed video game series depicts a fictional millennia-old struggle between the Order of Assassins, who fight for peace and free will, and the Knights Templar, who desire peace through order and control\\
......\\
    \vspace{1mm}
    
    \textit{Doc 2:}\\
    TITLE: Game of Throw-ins\\
CONTENT: Game of Throw-ins is a 2016 book by Irish author Paul Howard and is the sixteenth novel in the Ross O'Carroll-Kelly series.\\
The title refers to the TV series Game of Thrones and the rugby throw-in.\\
== Plot ==\\
Ross joins a struggling Seapoint rugby team. Ronan is in a turf war with a rival Love/Hate tour operator. Honor is in love with a Justin Bieber lookalike. Fionnuala is marrying a 92-year-old billionaire.\\
......\\
    \vspace{1mm}

    \textcolor[RGB]{158,160,161}{[N complete documents abbreviated]}\\
    \vspace{1mm}

    \textit{Doc N:}\\
    TITLE: Seedless in Seattle\\
CONTENT: Seedless in Seattle is a 2015 book by Irish author Paul Howard and is the fifteenth novel in the Ross O'Carroll-Kelly series.\\
The title refers to the 1993 film Sleepless in Seattle.\\
== Plot ==\\
Ross' father is going to Argentina to find his missing daughter Erika. Ross is dealing with Fionn's new personality, making an enemy of his daughter, and when he gets caught writing "The Fuck-it List" it's the final straw for Sorcha. She insists that Ross gets a vasectomy.\\
......
\end{tcolorbox}

\begin{tcolorbox}[colback=white, colframe=gray!50!black, title=Question Type: \textbf{Hypothetical Scenario}]
    \textbf{Question:}
    
        Imagine a hypothetical scenario where a comprehensive history of philosophy, published in the mid-18th century and instrumental in shaping the views of Enlightenment thinkers, chose to emphasize an episode from 1693 where a celebrated English polymath suffered a nervous breakdown and sent wild, accusatory letters to his friend, a prominent fellow philosopher. If this influential text had successfully framed the polymath primarily as a figure of mental instability, which specific Latin phrase from the inscription on the sarcophagus of his monument, completed 13 years before the book's publication, would stand in the most direct and ironic contradiction to this historical portrayal?

    \vspace{2mm}
    \textbf{Answer:}
    Qui, animi vi prope divinÃ¢.

    \vspace{2mm}
    \textbf{Reasoning Chain:} (Johann Jacob Brucker)-[authored]-(Historia Critica Philosophiae)-[positioned as a central philosophical figure]-(Newton)-[sent accusatory letters to]-(John Locke)-[influenced]-(Isaac Newton)-[is monument by]-(Newton's monument)-[is featured on]-(sarcophagus)

    \tcblower
    \textbf{Documents:}
    
    \textit{Doc 1:}\\
    TITLE: Isaac Newton\\
CONTENT: Sir Isaac Newton (4 January [O.S. 25 December] 1643 – 31 March [O.S. 20 March] 1727) was an English polymath active as a mathematician, physicist, astronomer, alchemist, theologian, and author. Newton was a key figure in the Scientific Revolution and the Enlightenment that followed. His book Philosophiæ Naturalis Principia Mathematica (Mathematical Principles of Natural Philosophy), first published in 1687, achieved the first great unification in physics and established classical mechanics. Newton also made seminal contributions to optics, and shares credit with German mathematician Gottfried Wilhelm Leibniz for formulating infinitesimal calculus, though he developed calculus years before Leibniz. Newton contributed to and refined the scientific method, and his work is considered the most influential in bringing forth modern science.\\
......\\
    \vspace{1mm}
    
    \textit{Doc 2:}\\
    TITLE: Francis Ronalds\\
CONTENT: Sir Francis Ronalds FRS (21 February 1788 – 8 August 1873) was an English scientist and inventor, and arguably the first electrical engineer. He was knighted for creating the first working electric telegraph over a substantial distance. In 1816 he laid an 8-mile (13 km) length of iron wire between wooden frames in his mother's garden and sent pulses using electrostatic generators. He also is known for creating the first electric clock in 1814.\\
......\\
    \vspace{1mm}

    \textcolor[RGB]{158,160,161}{[N complete documents abbreviated]}\\
    \vspace{1mm}

    \textit{Doc N:}\\
    TITLE: George Adams (scientist, died 1795)\\
CONTENT: George Adams the younger (1750–1795) was an English scientist, optician and scientific writer. He was mathematical instrument maker to King George III of Great Britain, succeeding his father George Adams in the post. He also made globes.\\
Around 1770, Adams invented the lucernal microscope, a type of projection microscope where the image is projected on a screen by a large oil lamp, as to make it easier to draw or trace the image.\\
In politics Adams was a Tory, and as such was received with favour at court by George III. He died 14 August 1795, at Southampton, and was succeeded in his business and in the post of mathematical instrument maker to the king by his brother, Dudley Adams.\\
......\\
\end{tcolorbox}

\begin{tcolorbox}[colback=white, colframe=gray!50!black, title=Question Type: \textbf{Temporal Reasoning}]
    \textbf{Question:}
    
         Calculate the number of years that passed between the year a wrestler, known for a lengthy WrestleMania winning streak, cost an opponent a championship match in retaliation for interference the previous month, and the year he first faced an opponent at that same flagship event who, a decade later, would vow to break his streak. If you add this duration to the year of that initial match, how many years prior was this calculated year to the first and only instance of an NFL team completing a perfect regular season under the 16-game schedule?

    \vspace{2mm}
    \textbf{Answer:} 1 Year.

    \vspace{2mm}
    \textbf{Reasoning Chain:} (Diesel)-[had match interfered in by]-(The Undertaker)-[had match against]-(Triple H)-[vowed to end]-(The Streak)-[took place at]-(WrestleMania)

    \tcblower
    \textbf{Documents:}
    
    \textit{Doc 1:}\\
    TITLE: Perfect season\\
CONTENT: A perfect season is a sports season, including any requisite playoff portion, in which a team remains and finishes undefeated and untied. The feat is extremely rare at the professional level of any team sport, but has occurred more commonly at the collegiate and scholastic levels in the United States. A perfect regular season (known by other names outside the United States) is a season excluding any playoffs, where a team remains undefeated and untied; it is less rare than a complete perfect season but still exceptional.\\
A perfect season may be part of a multi-season winning streak, or even a streak of perfect seasons.\\
Exhibition games are generally not counted toward standings, for or against. For example, the 1972 Miami Dolphins (below) lost three of their preseason ("exhibition" games in 1972 NFL vernacular) games but are considered to have had a perfect season.\\
......\\
    \vspace{1mm}
    
    \textit{Doc 2:}\\
    TITLE: Goldberg win streak\\
CONTENT: The Goldberg win streak was a lengthy series of victories that established the character of American professional wrestler Goldberg, following his debut on WCW Monday Nitro on September 22, 1997. The unprecedented win streak proved to be essential in making Goldberg the breakout star of World Championship Wrestling (WCW), propelling the rookie wrestler to main event status within a year of his first match, and would become a tool used by other promotions to build young stars into main event players.\\
......\\
    \vspace{1mm}

    \textcolor[RGB]{158,160,161}{[N complete documents abbreviated]}\\
    \vspace{1mm}

    \textit{Doc N:}\\
TITLE: The Streak (professional wrestling)\\
CONTENT: The Streak was a series of 21 consecutive victories for professional wrestler The Undertaker (Mark Calaway) at WWE's annual flagship marquee event, WrestleMania. It began at WrestleMania VII in 1991 when he beat Jimmy Snuka, with the final win coming against CM Punk at WrestleMania 29 in 2013; the Undertaker was absent from WrestleMania X in 1994 and WrestleMania 2000, owing to injury. Overall, he defeated 18 men during the Streak, which included three bouts with Triple H and two each opposite Kane and Shawn Michaels, as well as a handicap match against A-Train and Big Show at WrestleMania XIX.\\
The Streak became the cornerstone of WrestleMania, with a potential win over The Undertaker at the event being described as a greater honor than winning the WWE Championship. For years, debate had revolved around who, if anybody, should break the Streak, with prominent wrestlers giving comment. At WrestleMania XXX in 2014, The Undertaker lost by pinfall to Brock Lesnar, thus ending the Streak.\\
......
    
\end{tcolorbox}

\begin{tcolorbox}[colback=white, colframe=gray!50!black, title=Question Type: \textbf{Causal Analysis}]
    \textbf{Question:}
    
         A prominent English physician born in Kent in the 1570s, though not a member himself, defended a mystical movement whose manifestos circulated widely in Europe in the early 17th century. This movement\'s ideas are believed to have influenced a speculative fraternal organization that later promoted the scientific views of an English philosopher. This philosopher, in turn, openly criticized a Swiss physician from the German Renaissance, who is also known as the "father of toxicology." What specific, innovative medical practice, which represented a significant departure from the prevailing humoral theory, was a key contribution of this criticized Swiss physician?

    \vspace{2mm}
    \textbf{Answer:}
    Clinical diagnosis and the administration of highly specific medicines.
    
    \vspace{2mm}
    \textbf{Reasoning Chain:} (Urszula Szulakowska)-[authored article on]-(Utriusque Cosmi maioris scilicet et minoris metaphysica..)-[is magnum opus of]-(Fludd)-[was not a]-(Rosicrucian)-[influenced]-(Freemasonry)-[promoted the views of]-(Francis Bacon)-[vilified]-(Paracelsus)-[gave birth to]-(clinical diagnosis)

    \tcblower
    \textbf{Documents:}
    
    \textit{Doc 1:}\\
    TITLE: Paracelsus\\
CONTENT: Paracelsus (c. 1493 – 24 September 1541), born Theophrastus von Hohenheim (full name Philippus Aureolus Theophrastus Bombastus von Hohenheim), was a Swiss physician, alchemist, lay theologian, and philosopher of the German Renaissance.\\
He was a pioneer in several aspects of the "medical revolution" of the Renaissance, emphasizing the value of observation in combination with received wisdom. He is credited as the "father of toxicology". Paracelsus also had a substantial influence as a prophet or diviner, his "Prognostications" being studied by Rosicrucians in the 17th century. Paracelsianism is the early modern medical movement inspired by the study of his works.\\
......\\
    \vspace{1mm}
    
    \textit{Doc 2:}\\
    TITLE: Robert Fludd\\
CONTENT: Robert Fludd, also known as Robertus de Fluctibus (17 January 1574 – 8 September 1637), was a prominent English Paracelsian physician with both scientific and occult interests. He is remembered as an astrologer, mathematician, cosmologist, Qabalist, and Rosicrucian.\\
Fludd is best known for his compilations in occult philosophy. He had a celebrated exchange of views with Johannes Kepler concerning the scientific and hermetic approaches to knowledge.\\
== Early life ==\\
He was born at Milgate House, Bearsted, Kent, on 17 January 1573/4. He was the son of Sir Thomas Fludd, a high-ranking governmental official (Queen Elizabeth I's treasurer for war in Europe), and Member of Parliament. His mother was Elizabeth Andrews Fludd.  A collage of 12 Coats of Arms of Fludd ancestors are shown in the painting above his right shoulder. His paternal arms goes back to Rhirid Flaidd whose name originates from Welsh meaning bloody or red wolf.\\
......\\
    \vspace{1mm}

    \textcolor[RGB]{158,160,161}{[N complete documents abbreviated]}\\
    \vspace{1mm}

    \textit{Doc N:}\\
    TITLE: Robert Fludd\\
CONTENT: Robert Fludd, also known as Robertus de Fluctibus (17 January 1574 – 8 September 1637), was a prominent English Paracelsian physician with both scientific and occult interests. He is remembered as an astrologer, mathematician, cosmologist, Qabalist, and Rosicrucian.\\
Fludd is best known for his compilations in occult philosophy. He had a celebrated exchange of views with Johannes Kepler concerning the scientific and hermetic approaches to knowledge.\\
== Early life ==\\
He was born at Milgate House, Bearsted, Kent, on 17 January 1573/4. He was the son of Sir Thomas Fludd, a high-ranking governmental official (Queen Elizabeth I's treasurer for war in Europe), and Member of Parliament. His mother was Elizabeth Andrews Fludd.  A collage of 12 Coats of Arms of Fludd ancestors are shown in the painting above his right shoulder. His paternal arms goes back to Rhirid Flaidd whose name originates from Welsh meaning bloody or red wolf.\\
......
\end{tcolorbox}

\section{Case Study}

\subsection{General Performance Gain}
\label{appendix:aime_case}

As illustrated in the analysis of the AIME 2025 problem, QwenLong-L1.5-30B-A3B exhibits a marked improvement in reasoning capabilities compared to Qwen3-30B-A3B-Thinking, characterized by hypothesis refinement and strategic adaptation. 
Qwen3-30B-A3B-Thinking becomes entrenched in a qualitative calculus approach involving local extrema intervals. Failing to resolve the problem analytically, it ultimately resorts to heuristic approximations.
In contrast, QwenLong-L1.5-30B-A3B identifies a logical contradiction between the standard interpretation of ``local minima''-which typically implies continuous solution intervals-and the problem's explicit constraint requiring exactly three discrete values. This discrepancy triggers a hypothesis revision, leading the model to reinterpret the condition as a global minimum attained simultaneously at two distinct points (implying $f(a)=f(b)$). Consequently, the model shifts its strategy from graphical sign-chart analysis to algebraic synthesis. It constructs the polynomial identity $P(x) - mx = (x-a)^2(x-b)^2$, enabling an analytic solution for the parameters. This transition highlights the model's evolved capacity for self-correction and its ability to employ high-level algebraic modeling to satisfy complex geometric constraints.

\begin{tcolorbox}[colback=white, colframe=gray!50!black, title=\textbf{Case Study: Transition from Heuristic Guessing to Algebraic Synthesis}]
    
    \textbf{Question (AIME 2025 id29):}
    Let $f(x)=\frac{(x-18)(x-72)(x-98)(x-k)}{x}$. There exist exactly three positive real values of $k$ such that $f$ has a minimum at exactly two real values of $x$. Find the sum of these three values of $k$.
    
    \vspace{2mm}
    \textbf{Reference Answer:} 240
    
    \vspace{4mm}
    
    \renewcommand{\arraystretch}{1.5}
    \begin{tabularx}{\textwidth}{@{}p{3cm}X@{}}
        \toprule
        \textbf{Model} & \textbf{Reasoning Chain Summary} \\
        \midrule
        
        \textbf{Qwen3-30B-A3B-Thinking} \newline  & 
        Compute derivative $f'(x)$ \(\rightarrow\) Analyze critical points via $x g'(x) - g(x) = 0$ \(\rightarrow\) [...] \(\rightarrow\) Interpret "minimum" as local minima \(\rightarrow\) [\textcolor{red!70!black}{Confusion}] Realize local minima counts usually yield continuous intervals for $k$, contradicting "exactly three values" \(\rightarrow\) [\textcolor{red!70!black}{Heuristic Collapse}] Assume critical $k$ values correspond to the fixed roots of the numerator \(\rightarrow\) Select $k \in \{18, 72, 98\}$ \(\rightarrow\) Sum $= 188$. \newline \textbf{\textcolor{red!70!black}{Result: Incorrect}} \\
        
        \midrule
        
        \textbf{QwenLong-L1.5-30B-A3B}  & 
        Analyze critical points via $g(x)$ roots \(\rightarrow\) [\textcolor{green!60!black}{Meta-Cognitive Check}] Detect contradiction between "exactly three discrete values" and the interval nature of local minima counts \(\rightarrow\) [\textcolor{green!60!black}{Hypothesis Refinement}] Reinterpret condition as \textbf{Global Minimum} attained at two distinct points ($f(a)=f(b)$) \(\rightarrow\) [\textcolor{green!60!black}{Algebraic Synthesis}] Construct identity $P(x) - mx = (x-a)^2(x-b)^2$ \(\rightarrow\) Match coefficients to derive $(S_2 - p^2)^2 = 4S_4$ \(\rightarrow\) Solve quartic for $k$ \(\rightarrow\) Candidates $\{8, 32, 200, 512\}$ \(\rightarrow\) [\textcolor{green!60!black}{Verification}] Check signs of extrema; reject $k=512$ (one max, one min) \(\rightarrow\) Sum $= 8+32+200 = 240$. \newline \textbf{\textcolor{green!60!black}{Result: Correct}} \\
        \bottomrule
    \end{tabularx}
\end{tcolorbox}

\subsection{High entropy negative responses}
\begin{figure}
    \centering
    \begin{tcolorbox}[colback=white, colframe=gray!50!black, title=Case of a negative rollout in long-context RL for DocMath]\label{case:correct_step_in_negative_rollouts}
    To determine the total cash proceeds received from the sales of Eagle Crest, Three Hickory, and Centura land properties, and their percentage contribution to the net cash provided by investing activities in 2011, we\dots First, we need to identify the cash proceeds for each of the three sales:

    \begin{enumerate}
        \item \textbf{Eagle Crest Sale}: \dots
        \begin{itemize}
            \item Cash received = Sale price - Mortgage assumed = \$3.8M - \$2.4M = \textbf{\$1.4 million}
            \item This sale includes a storage warehouse (an income-producing property), so it is reported under \textbf{``Proceeds from sales of income-producing properties''} in the Consolidated Statements of Cash Flows.
        \end{itemize}

        \item \textbf{Three Hickory Land Sale}: \dots
        \begin{itemize}
            \item Cash received = Sale price (since no mortgage amount is mentioned, we assume the full \$1.3 million was received).
            \item This is a land-only sale, so it is reported under \textbf{``Proceeds from sales of land''}.
        \end{itemize}

        \item \textbf{Centura Land Sale}: \dots
        \begin{itemize}
            \item Cash received = Sale price - Mortgage assumed = \$13.0M - \$7.2M = \textbf{\$5.8 million}
            \item This is a land-only sale, so it is reported under \textbf{``Proceeds from sales of land''}. \textbf{Total cash proceeds} = \$1.4M (Eagle Crest) + \$1.3M (Three Hickory) + \$5.8M (Centura) = \textbf{\$8.5 million}.
        \end{itemize}
    \end{enumerate}
    \dots

    These figures represent the actual cash received from these categories of sales in the Cash Flow statement. \dots
    \begin{itemize}
        \item Three Hickory and Centura (land) = \$1,210 thousand
        \item Total = \$1,892 + \$1,210 = \textbf{\$3,102 thousand} (\$3.102 million).
    \end{itemize}
    The \textbf{net cash provided by investing activities} for 2011 is \textbf{\$2,128 thousand}. Now, to find the percentage of the net cash provided by investing activities attributable to these specific property sales:
    \textcolor{green!60!black}{\[
        \text{Percentage} = \left( \frac{3,102}{2,128} \right) \times 100 = 145.7\%
    \]}

    \textcolor{red}{This percentage exceeds 100\%, which is impossible. The error arises because the problem likely intended to ask for \dots Rounded to the nearest whole number, this is \textbf{55\%}.} \\
    \textcolor{red!60!black}{Therefore, the answer is 55.}
\end{tcolorbox}
    \caption{Example of a high-entropy negative rollout. This example illustrates a high-entropy negative response, where an initially correct reasoning path is abruptly invalidated by a final, erroneous self-correction, leading to an incorrect answer.}
    \label{fig:case_correct_steps_in_nega_steps}
\end{figure}

\begin{figure}
    \centering
    \begin{tcolorbox}[colback=white, colframe=gray!50!black, title=Case of a high-entropy negative rollout in long-context RL for Multi-Document QA]
    \textbf{Correct Response}\\
    \noindent\rule{\linewidth}{0.4pt}\\
    We are asked to compute a final year using a series of steps involving various dates, ages, counts, and events from the provided texts. Let's break it down step by step.\\
    \textbf{Step 1: Year of the first national anti-doping law in France}\\
    \dots\\
    \textbf{Step 2: Age of Danish cyclist Knud Enemark Jensen at his death}\\
    \dots\\
    \textbf{Step 3: Founding year of the World Drug-Free Powerlifting Federation (WDFPF)}\\
    \dots\\
    \textbf{Step 4: Add the number of months of Ilona Slupianek’s suspension}\\
    \dots\\
    \textbf{Step 5: From this year, subtract the number of years between the banning of boosting and the IPC's survey on its use at the 2008 Beijing Games}\\
    \dots\\
    \textbf{Step 6: Add the number of villains from the TV series ``The Games Affair''}\\
    \dots\\
    2004 + 3 = \textbf{2007}\\
    \dots\\
    \textbf{Step 7: Finally, subtract the album number of the comic ``Asterix at the Olympic Games''}\\
    \dots\\
    \textcolor{green!60!black}{Now subtract: 2007 - 12 = \textbf{1995}\\
   \textbf{ Therefore, the answer is (1995).}}\\\\

    \textbf{Incorrect Response}\\
    \noindent\rule{\linewidth}{0.4pt}\\
    We are asked to perform a series of calculations based on information from the provided articles. Let's go step by step, extracting the required data and performing the operations as described.\\
    \textbf{Step 1: Year of the first national anti-doping law in France}\\
    \dots\\
    \textbf{Step 2: Age of Danish cyclist Knud Enemark Jensen at his death}\\
    \dots\\
    \textbf{Step 3: Add these two numbers}\\
    \dots\\
    \textbf{Step 4: Identify the founding year of the World Drug-Free Powerlifting Federation (WDFPF)}\\
    \dots\\
    \textbf{Step 5: Add the number of months of Ilona Slupianek’s suspension}\\
    \dots\\
    \textbf{Step 6: Add the number of CIGEPS member states in 2015}\\
    \dots\\
    \textbf{Step 7: This gives the year the International Testing Agency was established}\\
    \textcolor{red!60!black}{subtract the album number of ``Asterix at the Olympic Games'' (12) $\rightarrow$ 2008 - 12 = \textbf{1996}\\
    \dots\\
    No --- it was \textbf{2007}. so if they of year 2007, and we as the about, near to 1996.\\
    \dots\\
    \textbf{Therefore, the answer is (1996)}}
    \end{tcolorbox}
    \caption{Example of a high-entropy negative rollout. An example of a high-entropy negative rollout where the reasoning process closely mirrors that of a correct solution. Towards the end of the process, the model successfully identifies a prior computational error but fails to rectify it, ultimately leading to an incorrect answer.}
    \label{fig:case_hig_entropy_2}
\end{figure}

\begin{figure}
    \centering
    \begin{tcolorbox}[colback=white, colframe=gray!50!black, title=Case of of a high entropy negative rollout in long-context RL for DocMath]
    We are asked to calculate the \textbf{estimated remaining useful life of the patents in years} as of \textbf{March 31, 2013}, using the following information:\\
    \ldots\\
\noindent\rule{\linewidth}{0.4pt}\\
\textbf{Step 1: Extract relevant data from the financial statements}\\
\ldots\\
\textbf{Step 2: Calculate total accumulated amortization as of March 31, 2013}\\
\ldots\\
\textbf{Step 3: Calculate total amortization expense since inception (Oct 1, 2007 -- Mar 31, 2013)}\\
\ldots\\
\textbf{Step 4: Determine the amortization rate per year}\\
\ldots\\
\textbf{Step 5: Estimate the total useful life (in years) of the patents}\\
\ldots\\
\textcolor{green!60!black}{for the future: 584{,}866 / 39{,}115.27 = 14.95 years.}
    
   \textcolor{red}{- on the basis of average amortization.\textbackslash n\textbackslash n- and because the rates for 2\textbackslash n\textbackslash n therefore, the answer is based on current rate of amortization is 38,788 per 6 months, though average is different, but for as a approximation as future amortization is to from the this and the  with of\textbackslash n\textbackslash n we and we \textbackslash n ( since you and to a of. wait. to other for the.\textbackslash n\textbackslash n- ** there is no to\textbackslash n\textbackslash n- “in Note 3” from the four the the.\textbackslash n\textbackslash n- the in the of if it the\textbackslash n\textbackslash n- you which one.\textbackslash n\textbackslash n- whether from to days.\textbackslash n\textbackslash n- event it is in the of  the and the it to is two days.\textbackslash n\textbackslash n- from after.\textbackslash n\textbackslash n- and the in only.\textbackslash n\textbackslash n. and later by.\textbackslash n\textbackslash n which so the\textbackslash n\textbackslash n- to being by\textbackslash n\textbackslash n. \textbackslash n. \textbackslash n. \textbackslash n- and how the. \textbackslash n- though.\textbackslash n\textbackslash n- maybe because the amortization as is a of of the.\textbackslash n\textbackslash n. but:\textbackslash n\textbackslash n see in in the of of  the over.\textbackslash n- but how to? must to the\textbackslash n\textbackslash n- the reduce.  better for \textbackslash n\textbackslash n- than above of, then.\textbackslash n\textbackslash n.\\
   ...\\
   \textbackslash n\textbackslash n we to to  wait.\textbackslash n\textbackslash n- you are to. \textbackslash n- case. for.\textbackslash n\textbackslash n- believe.\textbackslash n\textbackslash n- [ because so I final.\textbackslash n\textbackslash n- it 7.54 years. with\textbackslash n- try. \textbackslash n\textbackslash n- multiple time.\textbackslash n\textbackslash n- [ between \textbackslash n- [ and, to off\textbackslash n- make\textbackslash n- ( after reading the. \textbackslash n\textbackslash n- the correct and an and in the support because\textbackslash n\textbackslash n- all that\textbackslash n- the market industry data.\\
   \\
   \textcolor{red!60!black}{- Therefore, the answer is 7.54}}
    \end{tcolorbox}
    \caption{Example of a high-entropy negative rollout. The response exhibits frequent occurrences of ‘\textbackslash n\textbackslash n’ in the reasoning process after arriving at the correct answer, which degrades reasoning quality and ultimately leads to incorrect final answers.}
\end{figure}

\section{Prompt Template}

\definecolor{xmltag}{RGB}{0, 50, 120}  
\definecolor{variable}{RGB}{180, 40, 40} 

\begin{tcolorbox}[
    colback=gray!5!white,       
    colframe=gray!60!black,     
    title=\textbf{Prompt Template: Memory Agent Update}, 
    fonttitle=\bfseries,
    boxrule=0.4mm,
    sharp corners=south,        
    arc=3mm                     
]
\small
\newcommand{\tagfmt}[1]{\textcolor{xmltag}{\texttt{#1}}}
\newcommand{\varfmt}[1]{\textcolor{variable}{\texttt{#1}}}

\tagfmt{<problem>} \\
\varfmt{\{question\}} \\
\tagfmt{</problem>}

\vspace{0.5em}

\tagfmt{<previous\_memory>} \\
\varfmt{\{memory\}} \\
\tagfmt{</previous\_memory>}

\vspace{0.5em}

\tagfmt{<section>} \\
\varfmt{\{chunk\}} \\
\tagfmt{</section>}

\vspace{0.5em}

You are an assistant equipped with information memory capabilities. Based on the provided \tagfmt{<problem>}, \tagfmt{<previous\_memory>} (both \tagfmt{<memory>} and \tagfmt{<plan>}), and a \tagfmt{<section>} of a long article, you will traverse the sections of the article in the original order. 
Carefully read each section and update the memory with new information that helps answer the problem, while retaining all relevant details from the previous memory. 
After outputting the updated memory, do some simple planning to record the plan you need to answer the question, in addition to the existing memory.

\vspace{0.8em}
\textbf{[Important Notes]}
\begin{enumerate}[leftmargin=*, nosep, label=\arabic*.]
    \item You cannot see the full article; only focus on information memorization. You do not need to answer the \tagfmt{<problem>} directly at this stage.
    \item Ensure the memory is self-contained – meaning the final memory alone, without additional context, should contain sufficient information to fully answer the \tagfmt{<problem>}.
    \item Recognize that both the \tagfmt{<previous\_memory>.<memory>} and \tagfmt{<previous\_memory>.<plan>} may contain inaccuracies or be suboptimal for the current task. Therefore, 
    \begin{itemize}[leftmargin=1.2em]
        \item Pay close attention to the \tagfmt{<problem>}, critically evaluate the existing memory and plan, and make appropriate updates and adjustments to the memory based on the information found in the current \tagfmt{<section>}. 
        \item Your primary directive is to serve the \tagfmt{<problem>}, even if it means correcting prior memory or plans.
    \end{itemize}
    \item This is a strictly one-way reading process. You cannot go back to re-read previous sections. Therefore, 
    \begin{itemize}[leftmargin=1.2em]
        \item when you identify information to be extracted for the memory, you must directly and completely integrate the target content into the \tagfmt{<memory>} section. 
        \item Do not simply mark or reference key information; ensure the full, relevant segment is incorporated.
    \end{itemize}
    \item Carefully evaluate whether the extracted memory information truly serves the target requirement of the \tagfmt{<problem>}. 
    \begin{itemize}[leftmargin=1.2em]
        \item Avoid being misled by content that contains similar entities or semantic information but does not directly contribute to answering the \tagfmt{<problem>}. 
        \item Prioritize factual relevance and contribution to the overall solution.
    \end{itemize}
\end{enumerate}

\vspace{0.8em}
\textbf{[Output Format Example]} \\
Updated memory: \\
\tagfmt{<memory>} \\
(Your updated and consolidated memory here. Ensure it is self-contained and comprehensive for answering the \tagfmt{<problem>}.) \\
\tagfmt{</memory>}

\vspace{0.4em}

\tagfmt{<plan>} \\
(Your brief plan for the next stage of memory extraction, indicating what type of information you will look for in subsequent sections related to the \tagfmt{<problem>}.) \\
\tagfmt{</plan>} \\
\textbf{[End Example]}

\vspace{0.5em}
Updated memory:
\end{tcolorbox}

\definecolor{xmltag}{RGB}{0, 50, 120} 
\definecolor{variable}{RGB}{180, 40, 40}

\begin{tcolorbox}[
    colback=gray!5!white,       
    colframe=gray!60!black,     
    title=\textbf{Prompt Template: Memory Agent Final Answer}, 
    fonttitle=\bfseries,
    boxrule=0.4mm,
    sharp corners=south,        
    arc=3mm                     
]
\small
\newcommand{\tagfmt}[1]{\textcolor{xmltag}{\texttt{#1}}}
\newcommand{\varfmt}[1]{\textcolor{variable}{\texttt{#1}}}

You are presented with a problem and a previous memory. Please answer the problem based on the previous memory.

\vspace{0.5em}

\tagfmt{<problem>} \\
\varfmt{\{question\}} \\
\tagfmt{</problem>}

\vspace{0.5em}

\tagfmt{<memory>} \\
\varfmt{\{memory\}} \\
\tagfmt{</memory>}

\vspace{0.5em}

Your answer:
\end{tcolorbox}

\end{document}